%% file: main.tex
\documentclass[10pt,journal,compsoc]{IEEEtran}

\ifCLASSOPTIONcompsoc
  \usepackage[nocompress]{cite}
\else
  \usepackage{cite}
\fi

\ifCLASSINFOpdf
  \usepackage[pdftex]{graphicx}
\else
  \usepackage[dvips]{graphicx}
\fi

\usepackage{amsmath}
\usepackage{algorithmic}

\usepackage{array}

\ifCLASSOPTIONcompsoc
  \usepackage[caption=false,font=footnotesize,labelfont=sf,textfont=sf]{subfig}
\else
  \usepackage[caption=false,font=footnotesize]{subfig}
\fi

\usepackage{tabu}                      %
\usepackage{booktabs}                  %
\usepackage{color}
\definecolor{red}{rgb}{0.7,0,0}
\definecolor{green}{rgb}{0.0,0.7,0}
\definecolor{blue}{rgb}{0.00,0.00,0.75}
\definecolor{orange}{rgb}{0.72,0.22,0.06}
\definecolor{purple}{rgb}{0.6,0.0,0.6}
\definecolor{pink}{rgb}{1,0.03,0.5}
\definecolor{olive}{rgb}{0.4,0.6,0}
\definecolor{olive}{rgb}{0,0,0}

\usepackage[normalem]{ulem}
\usepackage{makecell}
\usepackage{multirow}
\usepackage{wasysym}
\usepackage[table,RGB]{xcolor}
\usepackage[T1]{fontenc}
\usepackage[maxfloats=256]{morefloats}
\begin{document}

\title{A Probabilistic Time-Evolving Approach to Scanpath Prediction}
\author{Daniel~Martin,
        Diego~Gutierrez,
        and Belen~Masia%
\IEEEcompsocitemizethanks{\IEEEcompsocthanksitem Daniel Martin, Diego Gutierrez, and Belen Masia are with Universidad de Zaragoza, I3A.\protect\\
\IEEEcompsocthanksitem Correspondence to: danims@unizar.es.}%
\thanks{Manuscript received April 2022.}
}

\markboth{A Probabilistic Time-Evolving Approach to Scanpath Prediction}%
{A Probabilistic Time-Evolving Approach to Scanpath Prediction}

\IEEEtitleabstractindextext{%
\begin{abstract}
Human visual attention is a complex phenomenon that has been studied for decades. Within it, the particular problem of scanpath prediction poses a challenge, particularly due to the inter- and intra-observer variability, among other reasons. Besides, most existing approaches to scanpath prediction have focused on optimizing the prediction of a gaze point given the previous ones. In this work, we present a probabilistic time-evolving approach to scanpath prediction, based on Bayesian deep learning. We optimize our model using a novel spatio-temporal loss function based on a combination of Kullback-Leibler divergence and dynamic time warping, jointly considering the spatial and temporal dimensions of scanpaths. Our scanpath prediction framework yields results that outperform those of current state-of-the-art approaches, and are almost on par with the human baseline, suggesting that our model is able to generate scanpaths whose behavior closely resembles those of the real ones. 
\end{abstract}

}

\maketitle
\IEEEdisplaynontitleabstractindextext
\IEEEpeerreviewmaketitle

\IEEEraisesectionheading{\section{Introduction}\label{sec:introduction}}

\IEEEPARstart{U}{nderstanding} human visual attention 
has been an active research area 
for decades. A plethora of works have been devoted to analyzing human attention when viewing content in different 
disciplines, including computer vision, graphics, neuroscience, or psychology.

\begin{figure*}[t!]
  \centering
  \includegraphics[width=\linewidth]{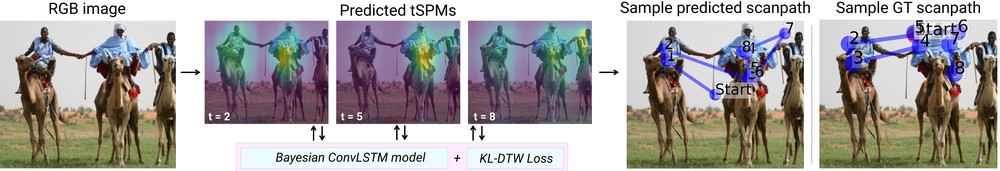}
  \caption{Overview of this work. We present a convolutional recurrent approach to scanpath prediction. Our model relies on Bayesian deep learning, and a novel spatialized representation of scanpaths. We propose a novel spatio-temporal loss function based on a combination of the Kullback-Leibler divergence and dynamic time warping, and predict time-evolving scanpath probabilistic maps (tSPM), well suited to the stochastic nature of human scanpaths. We evaluate our model and show how our generated scanpaths maintain the spatial and temporal characteristics of human scanpaths, while outperforming previous state-of-the-art approaches. An overview of the model itself is shown in Figure~\ref{fig:OurModel}.}
  \label{fig:teaser}
\end{figure*}

However, and regardless of the medium, gathering sufficiently large amounts of data to perform behavioral studies is a cumbersome and time-consuming task. Being able to generate virtual observers mimicking such attention process would greatly facilitate the process, thus helping achieve more significant advances in the fields. 

Many works have focused on, given an image, predicting where the attention of the observer is going to be directed to. 
Traditionally, the problem has been tackled through spatial, bottom-up analyses of the image, leading to the determination of salient areas represented as \emph{saliency maps}, which are  topographical representations by a scalar quantity of the conspicuity (i.e., saliency) at every location in the visual field~\cite{itti1998model, itti2001computational}.

 Although this may suffice for certain applications, saliency maps fail to capture the temporal dimension of gaze.
 This temporal information is relevant in a varied number of scenarios: How long does it take for an observer to find a specific object in an image? How should one design the layout of a 360º environment or scene? Will some distractor drive attention away from the main focal point, and if so, how and to what extent?  Current application areas where this temporal dimension prediction is relevant 
 range from marketing and product placement, webpage design or scene design, to analysis of visual pathologies or realistic eye motion simulation (e.g., for avatar animation).
 
To take into account this temporal information, a number of works have tackled the problem of \emph{scanpath} prediction~\cite{kummerer2021state}. A scanpath can be defined as a sequence of consecutive eye movements (i.e., fixations and saccades) through time and space~\cite{goldberg2010visual}. %
Gaze behavior is a complex phenomenon which  involves spatio-temporal dependencies~\cite{kapoula2021influence, martin2021scangan360},
 as well as a large inter- and intra-observer variability~\cite{le2016introducing, judd2009learning}. %

When attempting to model the temporal dimension of gaze, the problem is often posed as follows: given an input image $I$, and a sequence of gaze points $\{s_0,...,s_{t-1}\}$, the goal is to predict the next gaze point in the scanpath, $s_t$. Previous works either resort to heuristics and hand-crafted features~\cite{lemeur2015saccadic, tatler2009prominence}, or to data-driven methods~\cite{sun2019visual, bao2020human} to do this. %
However, most of the existing methods are designed to optimize the prediction of a \emph{single} fixation point, given the previous points; thus the scanpath is progressively built by concatenating successive single-point solutions. While this strategy is useful for several applications such as foveated rendering~\cite{arabadzhiyska2017saccade, nguyen2018your}, it may lead to increasing deviations from actual human viewing behavior and scanpath plausibility~\cite{fahimi2020metrics}.

In this paper we present a method to predict \textit{full, plausible} scanpaths 
given an input image $I$ (see Figure~\ref{fig:teaser}).
We leverage the fact that, despite the  inter- and intra-observer variability, common patterns and behaviors do emerge when humans observe certain content~\cite{ellis1985patterns}. This allows us to obtain not a single scanpath, but a \textit{distribution} of scanpaths within this common behavioral space. This distribution can then be sampled to generate individual scanpaths.

 We rely on convolutional long-short term memory networks (ConvLSTM)~\cite{xingjian2015convolutional}, since their recurrent architecture is well suited to capture the temporal dependency of each predicted point in a scanpath, while their convolutional nature has proven to be successful handling problems with both spatial and temporal dependencies. 
To obtain the distribution of plausible scanpaths we explicitly incorporate the inherent uncertainty of the problem into our model: Our ConvLSTM module is, for the first time, based on Bayesian deep learning, so that its weights are not deterministic, but sampled from a learned distribution instead. In addition, our network is trained using a novel spatio-temporal loss function that combines the benefits of the Kullback-Leibler divergence and dynamic time warping (DTW) for joint spatio-temporal optimization.

Our resulting trained model is able to generate a distribution of plausible scanpaths for a given input image, where each scanpath mimics the visual behavior of a human observer and takes less than one second to generate. We have validated our model both qualitatively and quantitatively, including an exhaustive set of existing metrics accounting for different scanpath characteristics~\cite{fahimi2020metrics}. Our model outperforms the state of the art, being almost on par with the human baseline.   %
We will make our code and model publicly available to encourage future research.

\input{sections/RelatedWork.tex}

\input{sections/Model.tex}

\input{sections/Evaluation.tex}
\input{sections/Conclusions.tex}

\ifCLASSOPTIONcompsoc
  \section*{Acknowledgments}
\else
  \section*{Acknowledgment}
\fi

This work has received funding from the European Research Council (ERC) under the European Union’s Horizon 2020 research and innovation programme (project CHAMELEON, Grant No 682080). This work has received funding from the European Union’s Horizon 2020 research and innovation programme under the Marie Skłodowska-Curie grant agreement No 956585. This project was also supported by a 2020 Leonardo Grant for Researchers and Cultural Creators, BBVA Foundation (the BBVA Foundation accepts no responsibility for the opinions, statements and contents included in the project and/or the results thereof, which are entirely the responsibility of the authors). This work has also received funding from Spain's Agencia Estatal de Investigación (project PID2019-105004GB-I00). Additionally, Daniel Martin was supported by a Gobierno de Aragon (2020-2024) predoctoral grant.

\ifCLASSOPTIONcaptionsoff
  \newpage
\fi

\bibliographystyle{IEEEtran}
\bibliography{bibliography}

\begin{IEEEbiography}[{\includegraphics[width=1in,height=1.25in,clip,keepaspectratio]{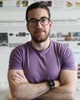}}]{Daniel Martin}
holds a MSc in Computer Science and is currently a PhD student at the Graphics \& Imaging Lab (Universidad de Zaragoza, Spain) under the supervision of Prof. Belen Masia and Prof. Diego Gutierrez. His research interests span modeling users’ behavior in multimodal environments, as well as studying how deep learning techniques can be used to achieve it. His work has been published at top venues, including ACM Transactions on Graphics, IEEE Transactions on Visualization and Computer Graphics, or ACM Computing Surveys, among others.
\end{IEEEbiography}

\begin{IEEEbiography}[{\includegraphics[width=1in,height=1.25in,clip,keepaspectratio]{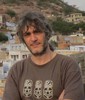}}]{Diego Gutierrez}
 is a Full Professor at Universidad de Zaragoza in Spain, where he leads the Graphics and Imaging Lab. His areas of research include physically-based rendering, virtual reality and computational imaging. He has published over 100 papers on those topics in top journals, including Nature. He has been Program Chair of several international conferences, including Eurographics in 2018 and the Eurographics Symposium on Rendering in 2012. He was Editor in Chief of ACM Transactions on Applied Perception from 2015 to 2017, and has served as Associate Editor in a number of other journals, including ACM Transactions on Graphics. He has received various awards, such as a Google Faculty Research Award. He received in 2016 an ERC Consolidator Grant.
\end{IEEEbiography}

\begin{IEEEbiography}[{\includegraphics[width=1in,height=1.25in,clip,keepaspectratio]{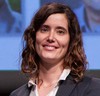}}]{Belen Masia} is a tenured Associate Professor in the Computer Science Department at Universidad de Zaragoza, Spain. She is a member of the Graphics \& Imaging Lab of the I3A Institute, and of the Vision, Image and Neurodevelopment Group of the IIS Aragon Institute. Her research focuses on the areas of computational imaging, applied perception, and virtual reality. Before, she was a postdoctoral researcher at Max Planck Institute for Informatics. Belen Masia is a Eurographics Junior Fellow. She is also the recipient of a Eurographics Young Researcher Award in 2017, a Eurographics PhD Award in 2015, an award to the top ten innovators below 35 in Spain from MIT Technology Review in 2014, and an NVIDIA Graduate Fellowship in 2012. She has served as an Associate Editor for ACM Transactions on Graphics, Computers and Graphics, and ACM Transactions on Applied Perception. She is also a co-founder of the startup DIVE Medical.
\end{IEEEbiography}

\newpage

\IEEEPARstart{W}{e} include here additional qualitative results from our model. Please refer to the main document for further analysis on quantitative and qualitative evaluations.

We show, for each image from our test set (see main document, Section 3.5):

\begin{itemize}
	\item Four predicted scanpaths with $th = 0.7$ (first row).
	\item Four predicted scanpaths with $th = 0.5$ (second row).
	\item Four predicted scanpaths with $th = 0.35$ (third row).
	\item The eight tSPM for one predicted scanpath (fourth and fifth row).
\end{itemize}

Further analysis on the effect of $th$ can be found in Section 4.3 in the main document, while the explanation on our tSPM can be found throughout Section 3 in the main document.

\subsection{Effect of image complexity}

\begin{figure*}[t!]
\centering
\includegraphics[width=\linewidth]{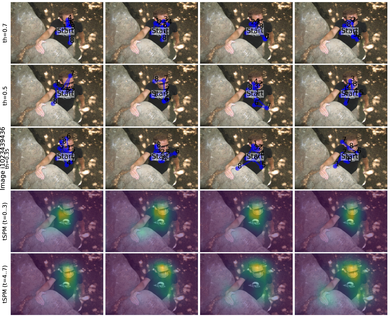}
\caption{Additional results for image \emph{i1023439436}. (First row) Four predicted scanpaths with $th=0.7$. (Second row) Four predicted scanpaths with $th=0.5$. (Third row) Four predicted scanpaths with $th=0.35$. (Fourth and fifth row) Predicted sequence of tSPM.}
\label{fig:0i1023439436}
\end{figure*}

\begin{figure*}[t!]
\centering
\includegraphics[width=\linewidth]{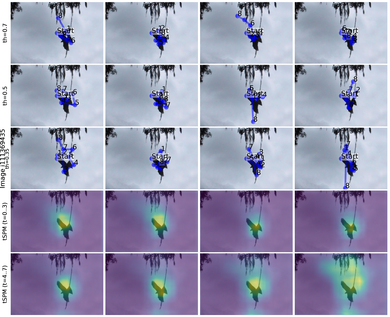}
\caption{Additional results for image \emph{i111369435}. (First row) Four predicted scanpaths with $th=0.7$. (Second row) Four predicted scanpaths with $th=0.5$. (Third row) Four predicted scanpaths with $th=0.35$. (Fourth and fifth row) Predicted sequence of tSPM.}
\label{fig:0i111369435}
\end{figure*}

\begin{figure*}[t!]
\centering
\includegraphics[width=\linewidth]{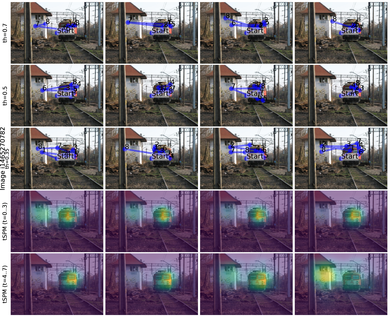}
\caption{Additional results for image \emph{i1465270782}. (First row) Four predicted scanpaths with $th=0.7$. (Second row) Four predicted scanpaths with $th=0.5$. (Third row) Four predicted scanpaths with $th=0.35$. (Fourth and fifth row) Predicted sequence of tSPM.}
\label{fig:0i1465270782}
\end{figure*}

\begin{figure*}[t!]
\centering
\includegraphics[width=\linewidth]{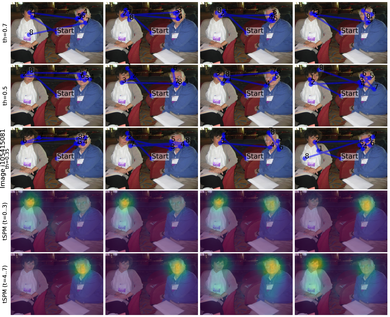}
\caption{Additional results for image \emph{i105415081}. (First row) Four predicted scanpaths with $th=0.7$. (Second row) Four predicted scanpaths with $th=0.5$. (Third row) Four predicted scanpaths with $th=0.35$. (Fourth and fifth row) Predicted sequence of tSPM.}
\label{fig:0i105415081}
\end{figure*}

\begin{figure*}[t!]
\centering
\includegraphics[width=\linewidth]{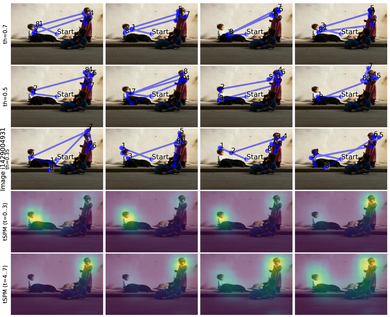}
\caption{Additional results for image \emph{i1429004931}. (First row) Four predicted scanpaths with $th=0.7$. (Second row) Four predicted scanpaths with $th=0.5$. (Third row) Four predicted scanpaths with $th=0.35$. (Fourth and fifth row) Predicted sequence of tSPM.}
\label{fig:0i1429004931}
\end{figure*}

\begin{figure*}[t!]
\centering
\includegraphics[width=\linewidth]{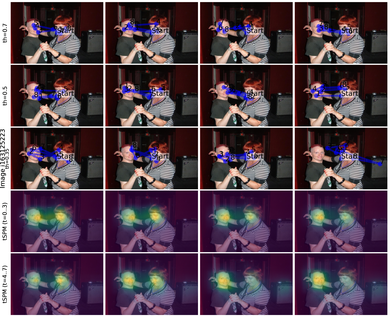}
\caption{Additional results for image \emph{i163125223}. (First row) Four predicted scanpaths with $th=0.7$. (Second row) Four predicted scanpaths with $th=0.5$. (Third row) Four predicted scanpaths with $th=0.35$. (Fourth and fifth row) Predicted sequence of tSPM.}
\label{fig:0i163125223}
\end{figure*}

\begin{figure*}[t!]
\centering
\includegraphics[width=\linewidth]{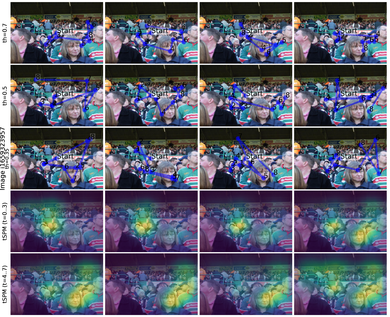}
\caption{Additional results for image \emph{i1659323957}. (First row) Four predicted scanpaths with $th=0.7$. (Second row) Four predicted scanpaths with $th=0.5$. (Third row) Four predicted scanpaths with $th=0.35$. (Fourth and fifth row) Predicted sequence of tSPM.}
\label{fig:0i1659323957}
\end{figure*}

\begin{figure*}[t!]
\centering
\includegraphics[width=\linewidth]{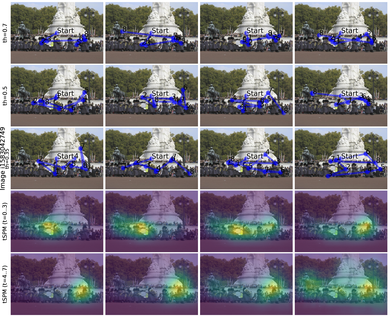}
\caption{Additional results for image \emph{i1583042749}. (First row) Four predicted scanpaths with $th=0.7$. (Second row) Four predicted scanpaths with $th=0.5$. (Third row) Four predicted scanpaths with $th=0.35$. (Fourth and fifth row) Predicted sequence of tSPM.}
\label{fig:0i1583042749}
\end{figure*}

\begin{figure*}[t!]
\centering
\includegraphics[width=\linewidth]{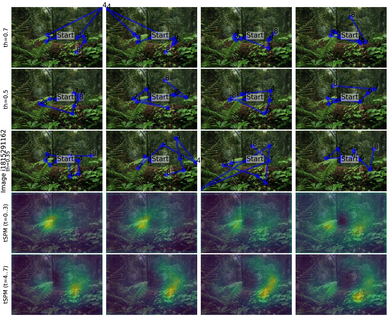}
\caption{Additional results for image \emph{i1815291162}. (First row) Four predicted scanpaths with $th=0.7$. (Second row) Four predicted scanpaths with $th=0.5$. (Third row) Four predicted scanpaths with $th=0.35$. (Fourth and fifth row) Predicted sequence of tSPM.}
\label{fig:0i1815291162}
\end{figure*}

\begin{figure*}[t!]
\centering
\includegraphics[width=\linewidth]{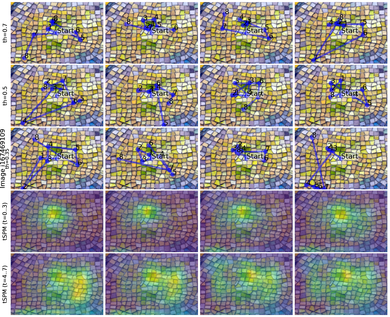}
\caption{Additional results for image \emph{i167469109}. (First row) Four predicted scanpaths with $th=0.7$. (Second row) Four predicted scanpaths with $th=0.5$. (Third row) Four predicted scanpaths with $th=0.35$. (Fourth and fifth row) Predicted sequence of tSPM.}
\label{fig:0i167469109}
\end{figure*}

As commented in the main document, the complexity of the image (i.e., the amount of regions of interest (ROI), or their absence) may affect the performance of our model. Thus, we show some particular examples of the behavior of our model in different circumstances, namely:

\begin{itemize}
	\item One ROI: Figures~\ref{fig:0i1023439436},~\ref{fig:0i111369435}, and~\ref{fig:0i1465270782}.
	\item Two ROI: Figures~\ref{fig:0i105415081},~\ref{fig:0i1429004931}, and~\ref{fig:0i163125223}.
	\item Multiple ROI: Figures~\ref{fig:0i1659323957}, and~\ref{fig:0i1583042749}.
	\item No clear ROI: Figures~\ref{fig:0i1815291162}, and~\ref{fig:0i167469109}.
\end{itemize}

\noindent while including the rest of the figures for completeness.

When few ROI are present, our model is able to correctly focus on them. However, when the image contains too many of them, our model is not always able to recognize them all, and focuses in a subset of them. When the image presents no ROI at all, our scanpaths become more erratic, as no clear interest zone can be focused.

\begin{figure*}[t!]
\centering
\includegraphics[width=\linewidth]{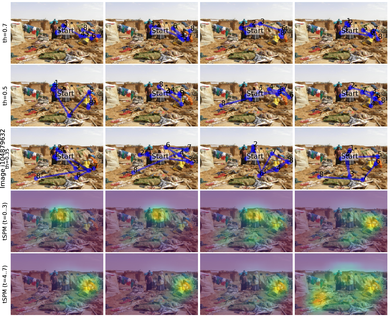}
\caption{Additional results for image \emph{i104879632}. (First row) Four predicted scanpaths with $th=0.7$. (Second row) Four predicted scanpaths with $th=0.5$. (Third row) Four predicted scanpaths with $th=0.35$. (Fourth and fifth row) Predicted sequence of tSPM.}
\label{fig:i104879632}
\end{figure*}

\begin{figure*}[t!]
\centering
\includegraphics[width=\linewidth]{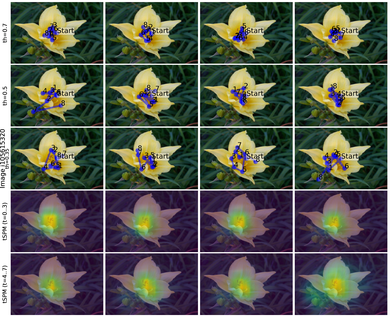}
\caption{Additional results for image \emph{i105615320}. (First row) Four predicted scanpaths with $th=0.7$. (Second row) Four predicted scanpaths with $th=0.5$. (Third row) Four predicted scanpaths with $th=0.35$. (Fourth and fifth row) Predicted sequence of tSPM.}
\label{fig:i105615320}
\end{figure*}

\begin{figure*}[t!]
\centering
\includegraphics[width=\linewidth]{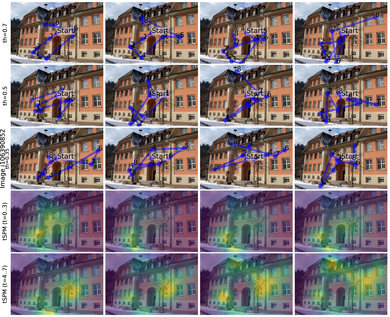}
\caption{Additional results for image \emph{i106390852}. (First row) Four predicted scanpaths with $th=0.7$. (Second row) Four predicted scanpaths with $th=0.5$. (Third row) Four predicted scanpaths with $th=0.35$. (Fourth and fifth row) Predicted sequence of tSPM.}
\label{fig:i106390852}
\end{figure*}

\begin{figure*}[t!]
\centering
\includegraphics[width=\linewidth]{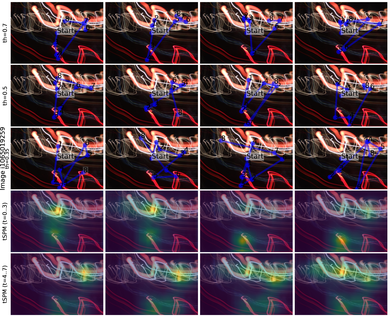}
\caption{Additional results for image \emph{i1065019259}. (First row) Four predicted scanpaths with $th=0.7$. (Second row) Four predicted scanpaths with $th=0.5$. (Third row) Four predicted scanpaths with $th=0.35$. (Fourth and fifth row) Predicted sequence of tSPM.}
\label{fig:i1065019259}
\end{figure*}

\begin{figure*}[t!]
\centering
\includegraphics[width=\linewidth]{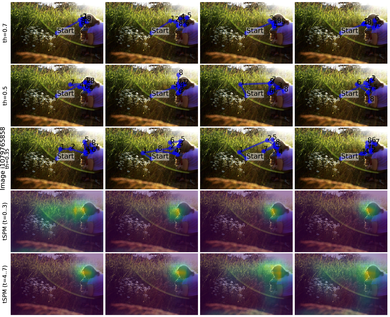}
\caption{Additional results for image \emph{i1079765858}. (First row) Four predicted scanpaths with $th=0.7$. (Second row) Four predicted scanpaths with $th=0.5$. (Third row) Four predicted scanpaths with $th=0.35$. (Fourth and fifth row) Predicted sequence of tSPM.}
\label{fig:i1079765858}
\end{figure*}

\begin{figure*}[t!]
\centering
\includegraphics[width=\linewidth]{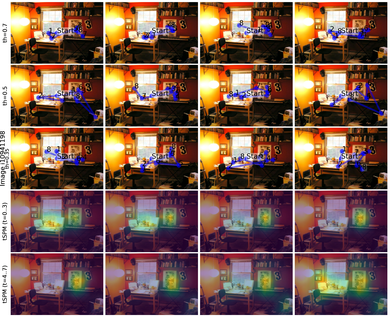}
\caption{Additional results for image \emph{i10941198}. (First row) Four predicted scanpaths with $th=0.7$. (Second row) Four predicted scanpaths with $th=0.5$. (Third row) Four predicted scanpaths with $th=0.35$. (Fourth and fifth row) Predicted sequence of tSPM.}
\label{fig:i10941198}
\end{figure*}

\begin{figure*}[t!]
\centering
\includegraphics[width=\linewidth]{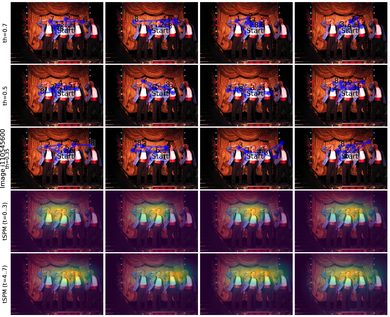}
\caption{Additional results for image \emph{i110545600}. (First row) Four predicted scanpaths with $th=0.7$. (Second row) Four predicted scanpaths with $th=0.5$. (Third row) Four predicted scanpaths with $th=0.35$. (Fourth and fifth row) Predicted sequence of tSPM.}
\label{fig:i110545600}
\end{figure*}

\begin{figure*}[t!]
\centering
\includegraphics[width=\linewidth]{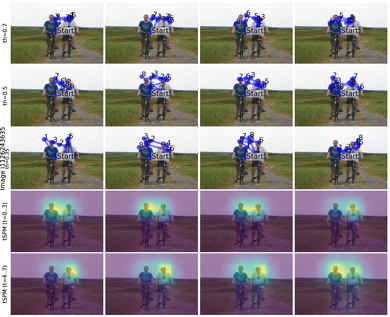}
\caption{Additional results for image \emph{i1126243635}. (First row) Four predicted scanpaths with $th=0.7$. (Second row) Four predicted scanpaths with $th=0.5$. (Third row) Four predicted scanpaths with $th=0.35$. (Fourth and fifth row) Predicted sequence of tSPM.}
\label{fig:i1126243635}
\end{figure*}

\begin{figure*}[t!]
\centering
\includegraphics[width=\linewidth]{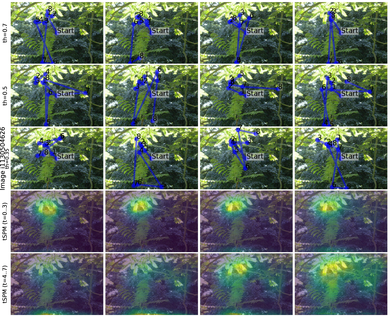}
\caption{Additional results for image \emph{i1130504626}. (First row) Four predicted scanpaths with $th=0.7$. (Second row) Four predicted scanpaths with $th=0.5$. (Third row) Four predicted scanpaths with $th=0.35$. (Fourth and fifth row) Predicted sequence of tSPM.}
\label{fig:i1130504626}
\end{figure*}

\begin{figure*}[t!]
\centering
\includegraphics[width=\linewidth]{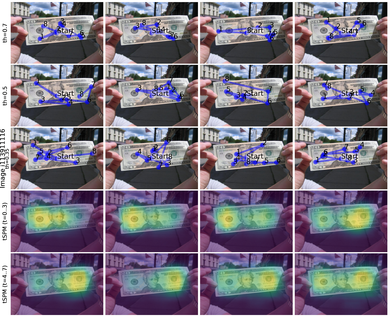}
\caption{Additional results for image \emph{i113911116}. (First row) Four predicted scanpaths with $th=0.7$. (Second row) Four predicted scanpaths with $th=0.5$. (Third row) Four predicted scanpaths with $th=0.35$. (Fourth and fifth row) Predicted sequence of tSPM.}
\label{fig:i113911116}
\end{figure*}

\begin{figure*}[t!]
\centering
\includegraphics[width=\linewidth]{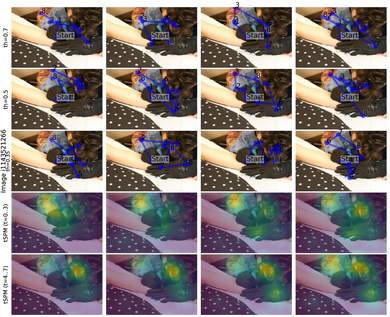}
\caption{Additional results for image \emph{i1143521266}. (First row) Four predicted scanpaths with $th=0.7$. (Second row) Four predicted scanpaths with $th=0.5$. (Third row) Four predicted scanpaths with $th=0.35$. (Fourth and fifth row) Predicted sequence of tSPM.}
\label{fig:i1143521266}
\end{figure*}

\begin{figure*}[t!]
\centering
\includegraphics[width=\linewidth]{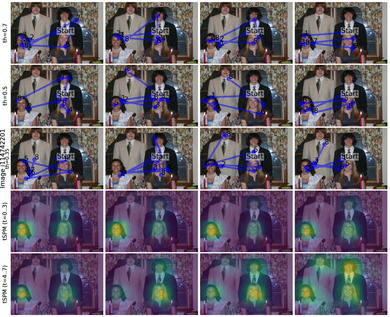}
\caption{Additional results for image \emph{i114742201}. (First row) Four predicted scanpaths with $th=0.7$. (Second row) Four predicted scanpaths with $th=0.5$. (Third row) Four predicted scanpaths with $th=0.35$. (Fourth and fifth row) Predicted sequence of tSPM.}
\label{fig:i114742201}
\end{figure*}

\begin{figure*}[t!]
\centering
\includegraphics[width=\linewidth]{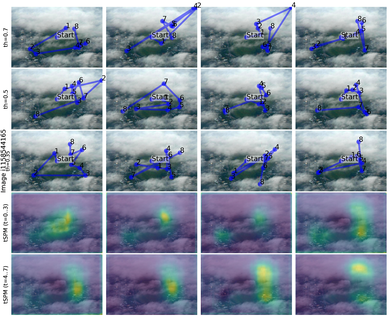}
\caption{Additional results for image \emph{i1158544165}. (First row) Four predicted scanpaths with $th=0.7$. (Second row) Four predicted scanpaths with $th=0.5$. (Third row) Four predicted scanpaths with $th=0.35$. (Fourth and fifth row) Predicted sequence of tSPM.}
\label{fig:i1158544165}
\end{figure*}

\begin{figure*}[t!]
\centering
\includegraphics[width=\linewidth]{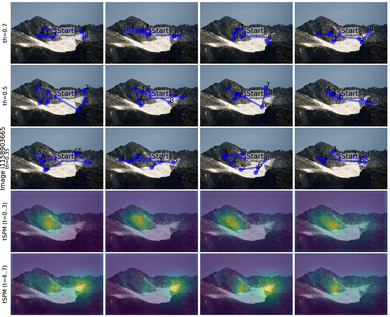}
\caption{Additional results for image \emph{i1158903665}. (First row) Four predicted scanpaths with $th=0.7$. (Second row) Four predicted scanpaths with $th=0.5$. (Third row) Four predicted scanpaths with $th=0.35$. (Fourth and fifth row) Predicted sequence of tSPM.}
\label{fig:i1158903665}
\end{figure*}

\begin{figure*}[t!]
\centering
\includegraphics[width=\linewidth]{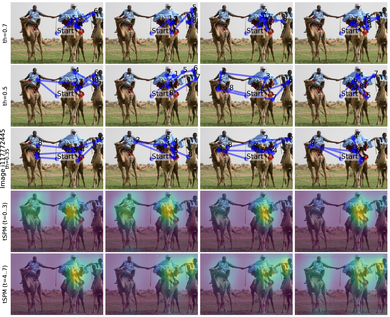}
\caption{Additional results for image \emph{i117772445}. (First row) Four predicted scanpaths with $th=0.7$. (Second row) Four predicted scanpaths with $th=0.5$. (Third row) Four predicted scanpaths with $th=0.35$. (Fourth and fifth row) Predicted sequence of tSPM.}
\label{fig:i117772445}
\end{figure*}

\begin{figure*}[t!]
\centering
\includegraphics[width=\linewidth]{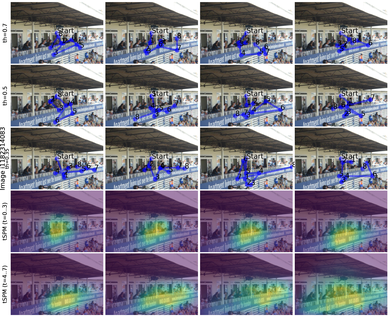}
\caption{Additional results for image \emph{i1182314083}. (First row) Four predicted scanpaths with $th=0.7$. (Second row) Four predicted scanpaths with $th=0.5$. (Third row) Four predicted scanpaths with $th=0.35$. (Fourth and fifth row) Predicted sequence of tSPM.}
\label{fig:i1182314083}
\end{figure*}

\begin{figure*}[t!]
\centering
\includegraphics[width=\linewidth]{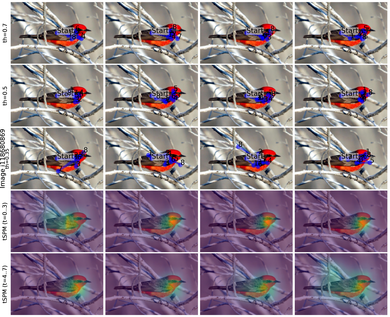}
\caption{Additional results for image \emph{i118680869}. (First row) Four predicted scanpaths with $th=0.7$. (Second row) Four predicted scanpaths with $th=0.5$. (Third row) Four predicted scanpaths with $th=0.35$. (Fourth and fifth row) Predicted sequence of tSPM.}
\label{fig:i118680869}
\end{figure*}

\begin{figure*}[t!]
\centering
\includegraphics[width=\linewidth]{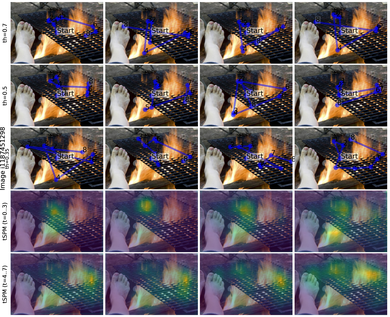}
\caption{Additional results for image \emph{i1187451298}. (First row) Four predicted scanpaths with $th=0.7$. (Second row) Four predicted scanpaths with $th=0.5$. (Third row) Four predicted scanpaths with $th=0.35$. (Fourth and fifth row) Predicted sequence of tSPM.}
\label{fig:i1187451298}
\end{figure*}

\begin{figure*}[t!]
\centering
\includegraphics[width=\linewidth]{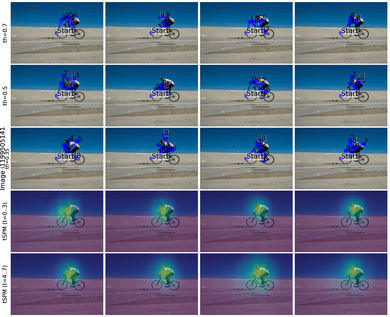}
\caption{Additional results for image \emph{i1199505141}. (First row) Four predicted scanpaths with $th=0.7$. (Second row) Four predicted scanpaths with $th=0.5$. (Third row) Four predicted scanpaths with $th=0.35$. (Fourth and fifth row) Predicted sequence of tSPM.}
\label{fig:i1199505141}
\end{figure*}

\begin{figure*}[t!]
\centering
\includegraphics[width=\linewidth]{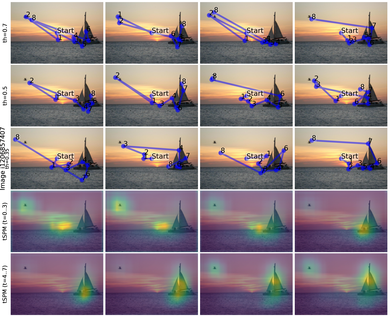}
\caption{Additional results for image \emph{i1206857407}. (First row) Four predicted scanpaths with $th=0.7$. (Second row) Four predicted scanpaths with $th=0.5$. (Third row) Four predicted scanpaths with $th=0.35$. (Fourth and fifth row) Predicted sequence of tSPM.}
\label{fig:i1206857407}
\end{figure*}

\begin{figure*}[t!]
\centering
\includegraphics[width=\linewidth]{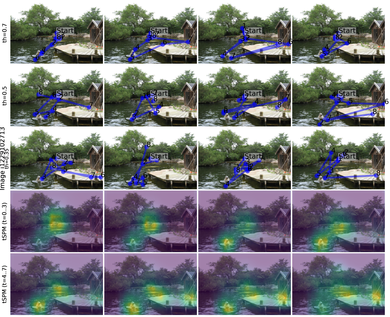}
\caption{Additional results for image \emph{i1229102713}. (First row) Four predicted scanpaths with $th=0.7$. (Second row) Four predicted scanpaths with $th=0.5$. (Third row) Four predicted scanpaths with $th=0.35$. (Fourth and fifth row) Predicted sequence of tSPM.}
\label{fig:i1229102713}
\end{figure*}

\begin{figure*}[t!]
\centering
\includegraphics[width=\linewidth]{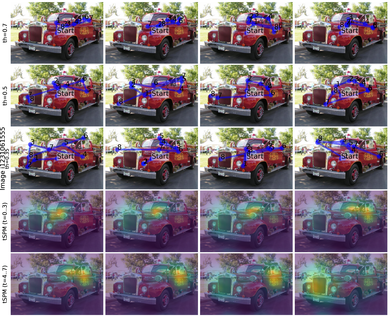}
\caption{Additional results for image \emph{i1231061555}. (First row) Four predicted scanpaths with $th=0.7$. (Second row) Four predicted scanpaths with $th=0.5$. (Third row) Four predicted scanpaths with $th=0.35$. (Fourth and fifth row) Predicted sequence of tSPM.}
\label{fig:i1231061555}
\end{figure*}

\begin{figure*}[t!]
\centering
\includegraphics[width=\linewidth]{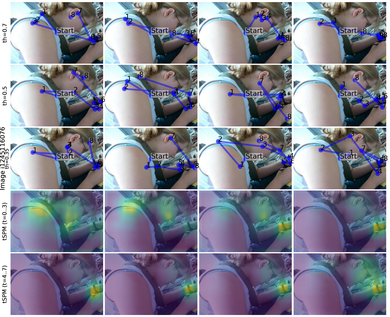}
\caption{Additional results for image \emph{i1245116076}. (First row) Four predicted scanpaths with $th=0.7$. (Second row) Four predicted scanpaths with $th=0.5$. (Third row) Four predicted scanpaths with $th=0.35$. (Fourth and fifth row) Predicted sequence of tSPM.}
\label{fig:i1245116076}
\end{figure*}

\begin{figure*}[t!]
\centering
\includegraphics[width=\linewidth]{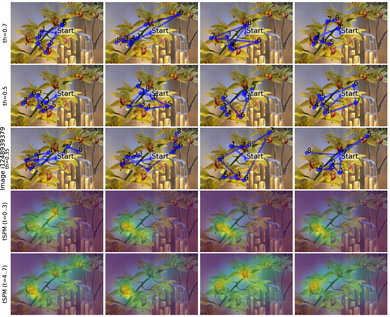}
\caption{Additional results for image \emph{i1248939379}. (First row) Four predicted scanpaths with $th=0.7$. (Second row) Four predicted scanpaths with $th=0.5$. (Third row) Four predicted scanpaths with $th=0.35$. (Fourth and fifth row) Predicted sequence of tSPM.}
\label{fig:i1248939379}
\end{figure*}

\begin{figure*}[t!]
\centering
\includegraphics[width=\linewidth]{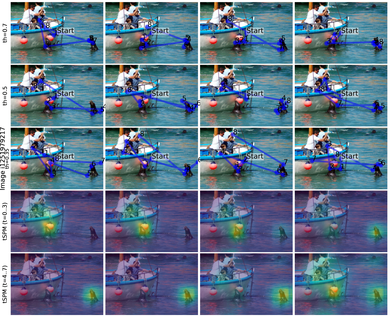}
\caption{Additional results for image \emph{i1251979217}. (First row) Four predicted scanpaths with $th=0.7$. (Second row) Four predicted scanpaths with $th=0.5$. (Third row) Four predicted scanpaths with $th=0.35$. (Fourth and fifth row) Predicted sequence of tSPM.}
\label{fig:i1251979217}
\end{figure*}

\begin{figure*}[t!]
\centering
\includegraphics[width=\linewidth]{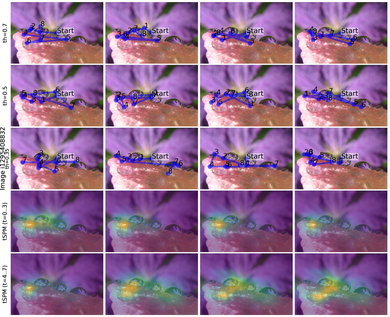}
\caption{Additional results for image \emph{i1295408832}. (First row) Four predicted scanpaths with $th=0.7$. (Second row) Four predicted scanpaths with $th=0.5$. (Third row) Four predicted scanpaths with $th=0.35$. (Fourth and fifth row) Predicted sequence of tSPM.}
\label{fig:i1295408832}
\end{figure*}

\begin{figure*}[t!]
\centering
\includegraphics[width=\linewidth]{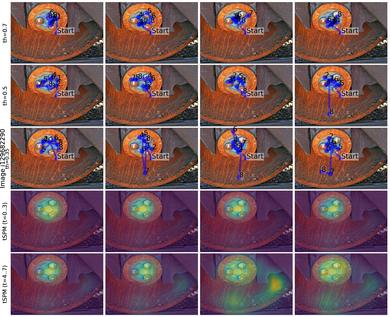}
\caption{Additional results for image \emph{i129682290}. (First row) Four predicted scanpaths with $th=0.7$. (Second row) Four predicted scanpaths with $th=0.5$. (Third row) Four predicted scanpaths with $th=0.35$. (Fourth and fifth row) Predicted sequence of tSPM.}
\label{fig:i129682290}
\end{figure*}

\begin{figure*}[t!]
\centering
\includegraphics[width=\linewidth]{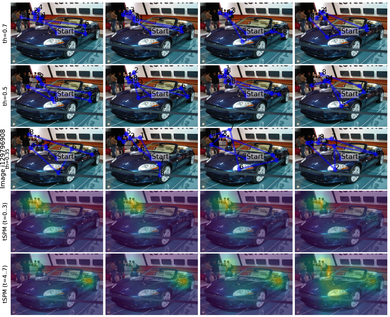}
\caption{Additional results for image \emph{i129796908}. (First row) Four predicted scanpaths with $th=0.7$. (Second row) Four predicted scanpaths with $th=0.5$. (Third row) Four predicted scanpaths with $th=0.35$. (Fourth and fifth row) Predicted sequence of tSPM.}
\label{fig:i129796908}
\end{figure*}

\begin{figure*}[t!]
\centering
\includegraphics[width=\linewidth]{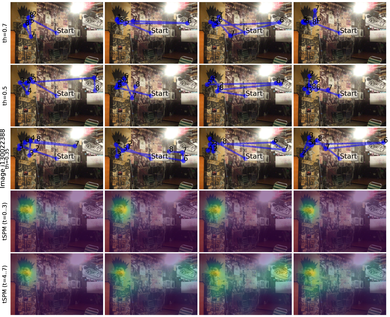}
\caption{Additional results for image \emph{i130022388}. (First row) Four predicted scanpaths with $th=0.7$. (Second row) Four predicted scanpaths with $th=0.5$. (Third row) Four predicted scanpaths with $th=0.35$. (Fourth and fifth row) Predicted sequence of tSPM.}
\label{fig:i130022388}
\end{figure*}

\begin{figure*}[t!]
\centering
\includegraphics[width=\linewidth]{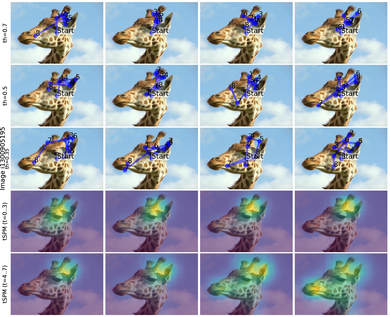}
\caption{Additional results for image \emph{i1300905195}. (First row) Four predicted scanpaths with $th=0.7$. (Second row) Four predicted scanpaths with $th=0.5$. (Third row) Four predicted scanpaths with $th=0.35$. (Fourth and fifth row) Predicted sequence of tSPM.}
\label{fig:i1300905195}
\end{figure*}

\begin{figure*}[t!]
\centering
\includegraphics[width=\linewidth]{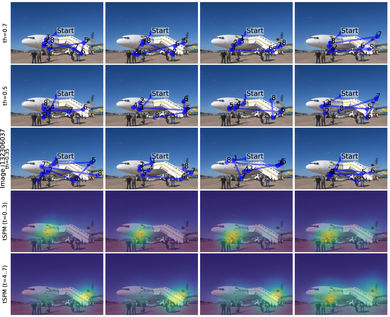}
\caption{Additional results for image \emph{i132306037}. (First row) Four predicted scanpaths with $th=0.7$. (Second row) Four predicted scanpaths with $th=0.5$. (Third row) Four predicted scanpaths with $th=0.35$. (Fourth and fifth row) Predicted sequence of tSPM.}
\label{fig:i132306037}
\end{figure*}

\begin{figure*}[t!]
\centering
\includegraphics[width=\linewidth]{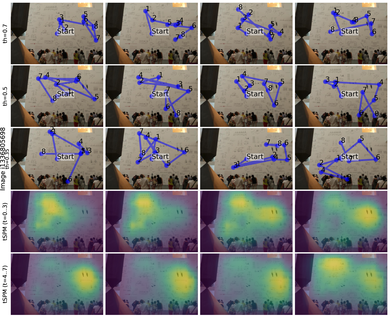}
\caption{Additional results for image \emph{i1336805698}. (First row) Four predicted scanpaths with $th=0.7$. (Second row) Four predicted scanpaths with $th=0.5$. (Third row) Four predicted scanpaths with $th=0.35$. (Fourth and fifth row) Predicted sequence of tSPM.}
\label{fig:i1336805698}
\end{figure*}

\begin{figure*}[t!]
\centering
\includegraphics[width=\linewidth]{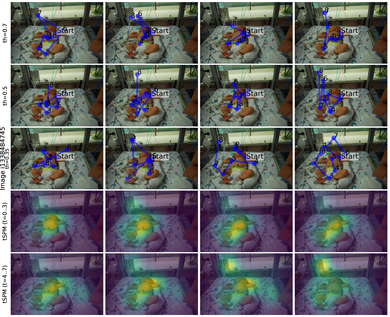}
\caption{Additional results for image \emph{i1338484745}. (First row) Four predicted scanpaths with $th=0.7$. (Second row) Four predicted scanpaths with $th=0.5$. (Third row) Four predicted scanpaths with $th=0.35$. (Fourth and fifth row) Predicted sequence of tSPM.}
\label{fig:i1338484745}
\end{figure*}

\begin{figure*}[t!]
\centering
\includegraphics[width=\linewidth]{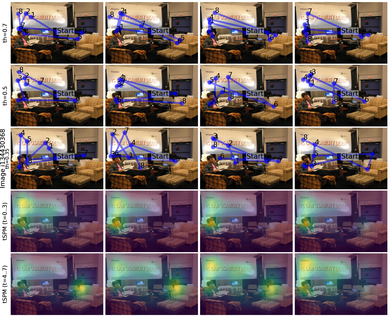}
\caption{Additional results for image \emph{i134430368}. (First row) Four predicted scanpaths with $th=0.7$. (Second row) Four predicted scanpaths with $th=0.5$. (Third row) Four predicted scanpaths with $th=0.35$. (Fourth and fifth row) Predicted sequence of tSPM.}
\label{fig:i134430368}
\end{figure*}

\begin{figure*}[t!]
\centering
\includegraphics[width=\linewidth]{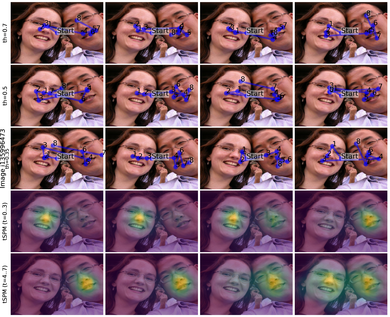}
\caption{Additional results for image \emph{i135996473}. (First row) Four predicted scanpaths with $th=0.7$. (Second row) Four predicted scanpaths with $th=0.5$. (Third row) Four predicted scanpaths with $th=0.35$. (Fourth and fifth row) Predicted sequence of tSPM.}
\label{fig:i135996473}
\end{figure*}

\begin{figure*}[t!]
\centering
\includegraphics[width=\linewidth]{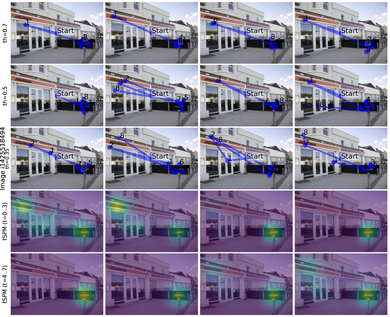}
\caption{Additional results for image \emph{i1425518494}. (First row) Four predicted scanpaths with $th=0.7$. (Second row) Four predicted scanpaths with $th=0.5$. (Third row) Four predicted scanpaths with $th=0.35$. (Fourth and fifth row) Predicted sequence of tSPM.}
\label{fig:i1425518494}
\end{figure*}

\begin{figure*}[t!]
\centering
\includegraphics[width=\linewidth]{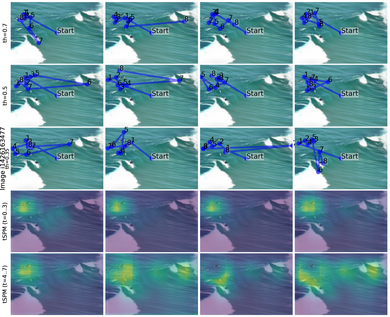}
\caption{Additional results for image \emph{i1426163477}. (First row) Four predicted scanpaths with $th=0.7$. (Second row) Four predicted scanpaths with $th=0.5$. (Third row) Four predicted scanpaths with $th=0.35$. (Fourth and fifth row) Predicted sequence of tSPM.}
\label{fig:i1426163477}
\end{figure*}

\begin{figure*}[t!]
\centering
\includegraphics[width=\linewidth]{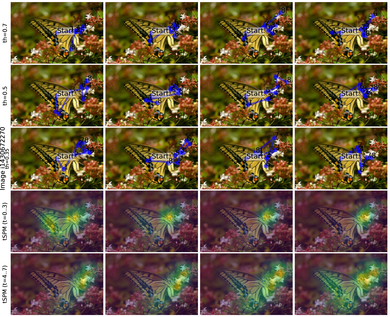}
\caption{Additional results for image \emph{i1430672270}. (First row) Four predicted scanpaths with $th=0.7$. (Second row) Four predicted scanpaths with $th=0.5$. (Third row) Four predicted scanpaths with $th=0.35$. (Fourth and fifth row) Predicted sequence of tSPM.}
\label{fig:i1430672270}
\end{figure*}

\begin{figure*}[t!]
\centering
\includegraphics[width=\linewidth]{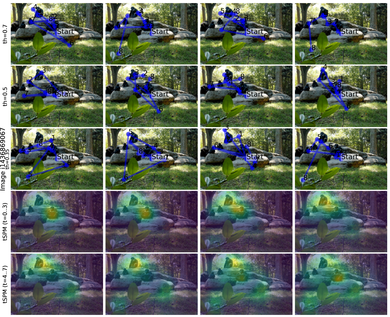}
\caption{Additional results for image \emph{i1436869067}. (First row) Four predicted scanpaths with $th=0.7$. (Second row) Four predicted scanpaths with $th=0.5$. (Third row) Four predicted scanpaths with $th=0.35$. (Fourth and fifth row) Predicted sequence of tSPM.}
\label{fig:i1436869067}
\end{figure*}

\begin{figure*}[t!]
\centering
\includegraphics[width=\linewidth]{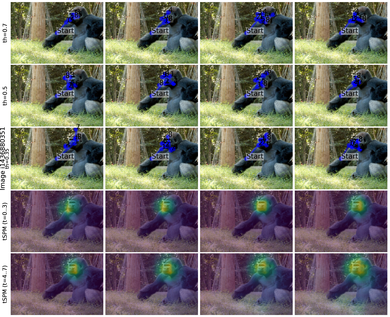}
\caption{Additional results for image \emph{i1436880351}. (First row) Four predicted scanpaths with $th=0.7$. (Second row) Four predicted scanpaths with $th=0.5$. (Third row) Four predicted scanpaths with $th=0.35$. (Fourth and fifth row) Predicted sequence of tSPM.}
\label{fig:i1436880351}
\end{figure*}

\begin{figure*}[t!]
\centering
\includegraphics[width=\linewidth]{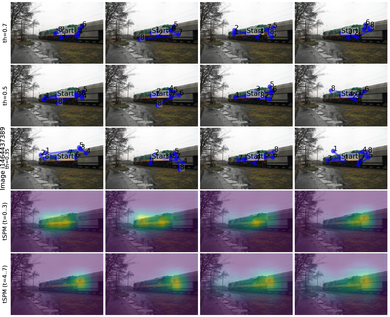}
\caption{Additional results for image \emph{i1464437389}. (First row) Four predicted scanpaths with $th=0.7$. (Second row) Four predicted scanpaths with $th=0.5$. (Third row) Four predicted scanpaths with $th=0.35$. (Fourth and fifth row) Predicted sequence of tSPM.}
\label{fig:i1464437389}
\end{figure*}

\begin{figure*}[t!]
\centering
\includegraphics[width=\linewidth]{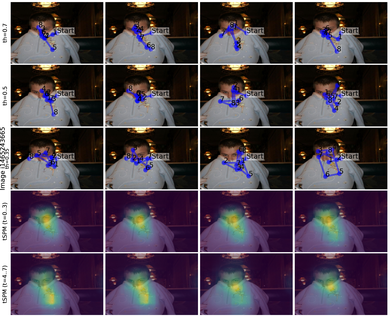}
\caption{Additional results for image \emph{i1465243665}. (First row) Four predicted scanpaths with $th=0.7$. (Second row) Four predicted scanpaths with $th=0.5$. (Third row) Four predicted scanpaths with $th=0.35$. (Fourth and fifth row) Predicted sequence of tSPM.}
\label{fig:i1465243665}
\end{figure*}

\begin{figure*}[t!]
\centering
\includegraphics[width=\linewidth]{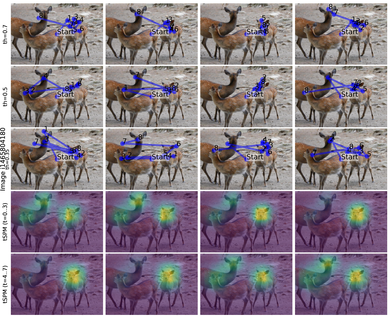}
\caption{Additional results for image \emph{i1465804180}. (First row) Four predicted scanpaths with $th=0.7$. (Second row) Four predicted scanpaths with $th=0.5$. (Third row) Four predicted scanpaths with $th=0.35$. (Fourth and fifth row) Predicted sequence of tSPM.}
\label{fig:i1465804180}
\end{figure*}

\begin{figure*}[t!]
\centering
\includegraphics[width=\linewidth]{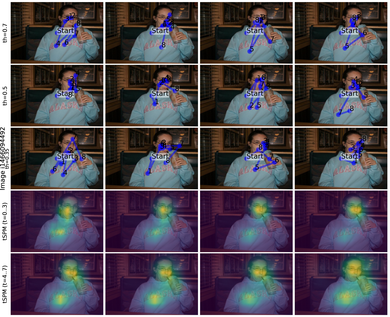}
\caption{Additional results for image \emph{i1466094492}. (First row) Four predicted scanpaths with $th=0.7$. (Second row) Four predicted scanpaths with $th=0.5$. (Third row) Four predicted scanpaths with $th=0.35$. (Fourth and fifth row) Predicted sequence of tSPM.}
\label{fig:i1466094492}
\end{figure*}

\begin{figure*}[t!]
\centering
\includegraphics[width=\linewidth]{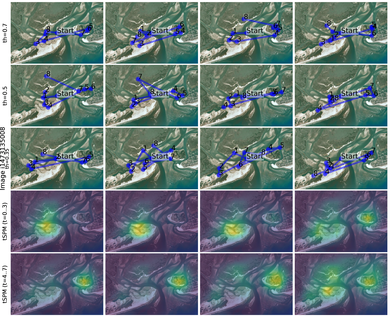}
\caption{Additional results for image \emph{i1473135008}. (First row) Four predicted scanpaths with $th=0.7$. (Second row) Four predicted scanpaths with $th=0.5$. (Third row) Four predicted scanpaths with $th=0.35$. (Fourth and fifth row) Predicted sequence of tSPM.}
\label{fig:i1473135008}
\end{figure*}

\begin{figure*}[t!]
\centering
\includegraphics[width=\linewidth]{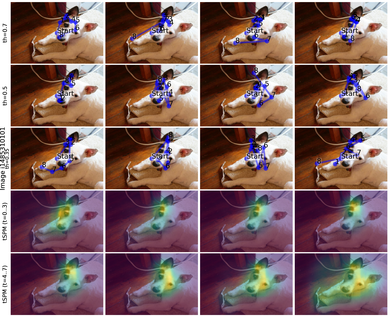}
\caption{Additional results for image \emph{i1485310101}. (First row) Four predicted scanpaths with $th=0.7$. (Second row) Four predicted scanpaths with $th=0.5$. (Third row) Four predicted scanpaths with $th=0.35$. (Fourth and fifth row) Predicted sequence of tSPM.}
\label{fig:i1485310101}
\end{figure*}

\begin{figure*}[t!]
\centering
\includegraphics[width=\linewidth]{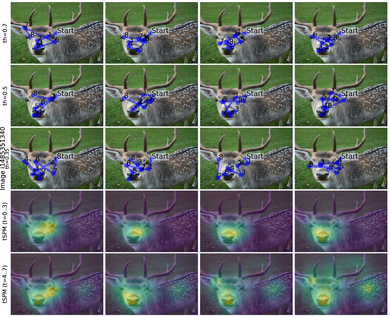}
\caption{Additional results for image \emph{i1485351340}. (First row) Four predicted scanpaths with $th=0.7$. (Second row) Four predicted scanpaths with $th=0.5$. (Third row) Four predicted scanpaths with $th=0.35$. (Fourth and fifth row) Predicted sequence of tSPM.}
\label{fig:i1485351340}
\end{figure*}

\begin{figure*}[t!]
\centering
\includegraphics[width=\linewidth]{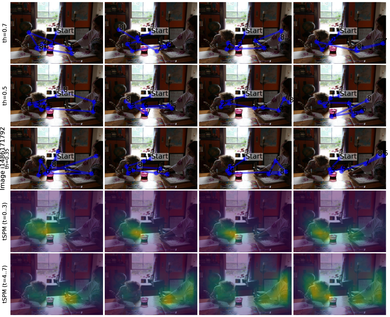}
\caption{Additional results for image \emph{i1486171792}. (First row) Four predicted scanpaths with $th=0.7$. (Second row) Four predicted scanpaths with $th=0.5$. (Third row) Four predicted scanpaths with $th=0.35$. (Fourth and fifth row) Predicted sequence of tSPM.}
\label{fig:i1486171792}
\end{figure*}

\begin{figure*}[t!]
\centering
\includegraphics[width=\linewidth]{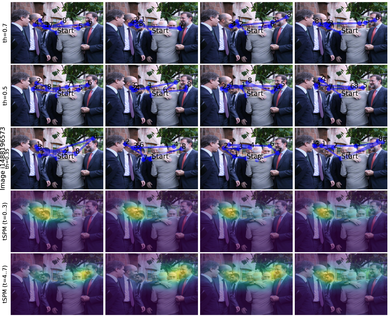}
\caption{Additional results for image \emph{i1488196573}. (First row) Four predicted scanpaths with $th=0.7$. (Second row) Four predicted scanpaths with $th=0.5$. (Third row) Four predicted scanpaths with $th=0.35$. (Fourth and fifth row) Predicted sequence of tSPM.}
\label{fig:i1488196573}
\end{figure*}

\begin{figure*}[t!]
\centering
\includegraphics[width=\linewidth]{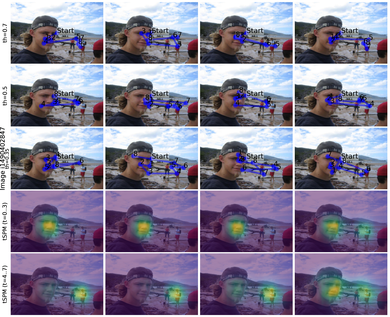}
\caption{Additional results for image \emph{i1490402847}. (First row) Four predicted scanpaths with $th=0.7$. (Second row) Four predicted scanpaths with $th=0.5$. (Third row) Four predicted scanpaths with $th=0.35$. (Fourth and fifth row) Predicted sequence of tSPM.}
\label{fig:i1490402847}
\end{figure*}

\begin{figure*}[t!]
\centering
\includegraphics[width=\linewidth]{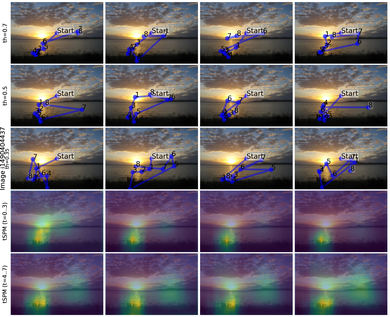}
\caption{Additional results for image \emph{i1490404437}. (First row) Four predicted scanpaths with $th=0.7$. (Second row) Four predicted scanpaths with $th=0.5$. (Third row) Four predicted scanpaths with $th=0.35$. (Fourth and fifth row) Predicted sequence of tSPM.}
\label{fig:i1490404437}
\end{figure*}

\begin{figure*}[t!]
\centering
\includegraphics[width=\linewidth]{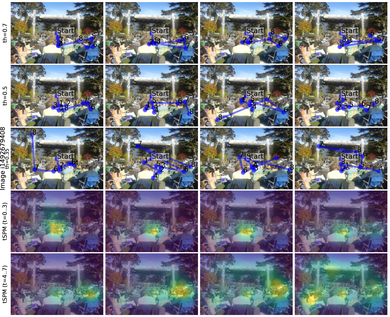}
\caption{Additional results for image \emph{i1492679408}. (First row) Four predicted scanpaths with $th=0.7$. (Second row) Four predicted scanpaths with $th=0.5$. (Third row) Four predicted scanpaths with $th=0.35$. (Fourth and fifth row) Predicted sequence of tSPM.}
\label{fig:i1492679408}
\end{figure*}

\begin{figure*}[t!]
\centering
\includegraphics[width=\linewidth]{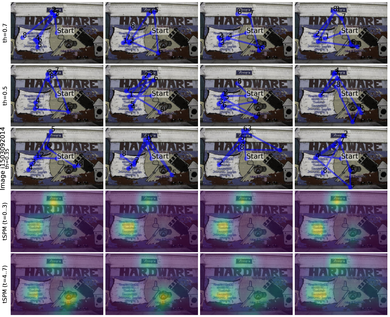}
\caption{Additional results for image \emph{i1503092014}. (First row) Four predicted scanpaths with $th=0.7$. (Second row) Four predicted scanpaths with $th=0.5$. (Third row) Four predicted scanpaths with $th=0.35$. (Fourth and fifth row) Predicted sequence of tSPM.}
\label{fig:i1503092014}
\end{figure*}

\begin{figure*}[t!]
\centering
\includegraphics[width=\linewidth]{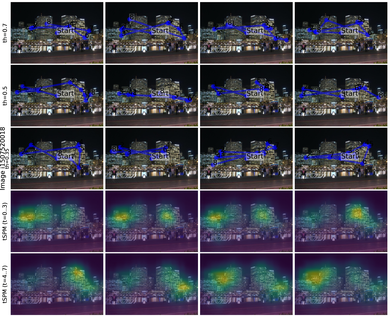}
\caption{Additional results for image \emph{i1507520018}. (First row) Four predicted scanpaths with $th=0.7$. (Second row) Four predicted scanpaths with $th=0.5$. (Third row) Four predicted scanpaths with $th=0.35$. (Fourth and fifth row) Predicted sequence of tSPM.}
\label{fig:i1507520018}
\end{figure*}

\begin{figure*}[t!]
\centering
\includegraphics[width=\linewidth]{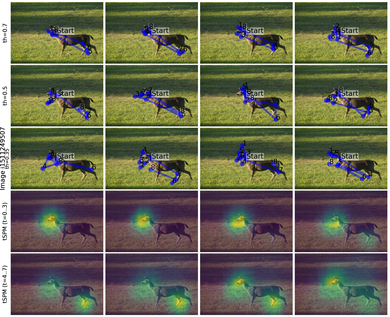}
\caption{Additional results for image \emph{i1511249507}. (First row) Four predicted scanpaths with $th=0.7$. (Second row) Four predicted scanpaths with $th=0.5$. (Third row) Four predicted scanpaths with $th=0.35$. (Fourth and fifth row) Predicted sequence of tSPM.}
\label{fig:i1511249507}
\end{figure*}

\begin{figure*}[t!]
\centering
\includegraphics[width=\linewidth]{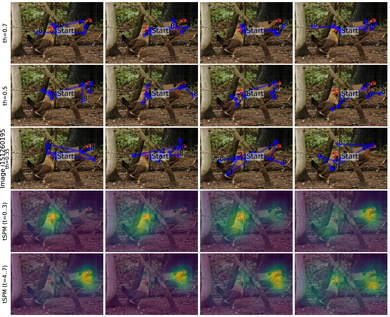}
\caption{Additional results for image \emph{i151260195}. (First row) Four predicted scanpaths with $th=0.7$. (Second row) Four predicted scanpaths with $th=0.5$. (Third row) Four predicted scanpaths with $th=0.35$. (Fourth and fifth row) Predicted sequence of tSPM.}
\label{fig:i151260195}
\end{figure*}

\begin{figure*}[t!]
\centering
\includegraphics[width=\linewidth]{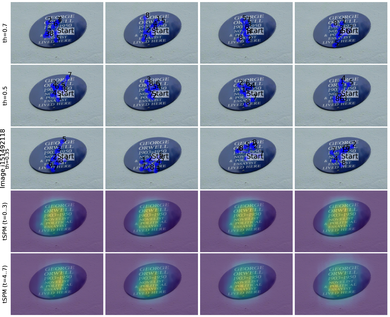}
\caption{Additional results for image \emph{i151492118}. (First row) Four predicted scanpaths with $th=0.7$. (Second row) Four predicted scanpaths with $th=0.5$. (Third row) Four predicted scanpaths with $th=0.35$. (Fourth and fifth row) Predicted sequence of tSPM.}
\label{fig:i151492118}
\end{figure*}

\begin{figure*}[t!]
\centering
\includegraphics[width=\linewidth]{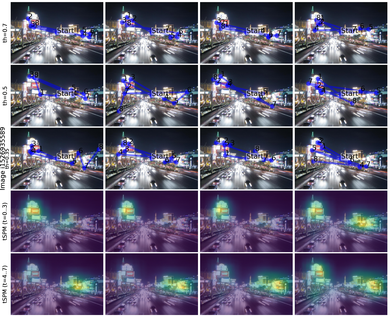}
\caption{Additional results for image \emph{i1526935589}. (First row) Four predicted scanpaths with $th=0.7$. (Second row) Four predicted scanpaths with $th=0.5$. (Third row) Four predicted scanpaths with $th=0.35$. (Fourth and fifth row) Predicted sequence of tSPM.}
\label{fig:i1526935589}
\end{figure*}

\begin{figure*}[t!]
\centering
\includegraphics[width=\linewidth]{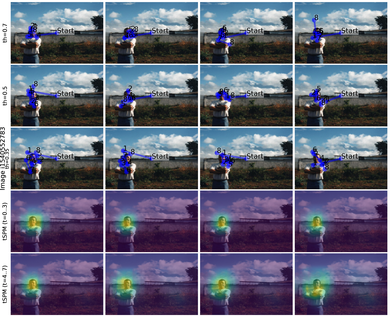}
\caption{Additional results for image \emph{i1540552783}. (First row) Four predicted scanpaths with $th=0.7$. (Second row) Four predicted scanpaths with $th=0.5$. (Third row) Four predicted scanpaths with $th=0.35$. (Fourth and fifth row) Predicted sequence of tSPM.}
\label{fig:i1540552783}
\end{figure*}

\begin{figure*}[t!]
\centering
\includegraphics[width=\linewidth]{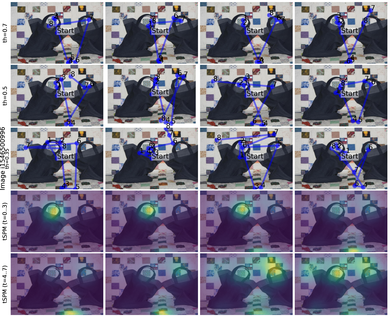}
\caption{Additional results for image \emph{i1546500996}. (First row) Four predicted scanpaths with $th=0.7$. (Second row) Four predicted scanpaths with $th=0.5$. (Third row) Four predicted scanpaths with $th=0.35$. (Fourth and fifth row) Predicted sequence of tSPM.}
\label{fig:i1546500996}
\end{figure*}

\begin{figure*}[t!]
\centering
\includegraphics[width=\linewidth]{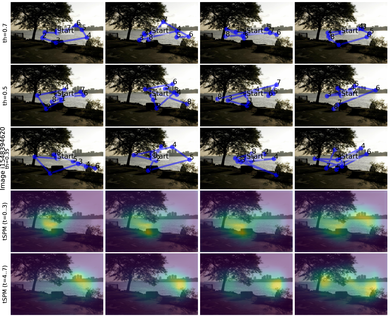}
\caption{Additional results for image \emph{i1548394620}. (First row) Four predicted scanpaths with $th=0.7$. (Second row) Four predicted scanpaths with $th=0.5$. (Third row) Four predicted scanpaths with $th=0.35$. (Fourth and fifth row) Predicted sequence of tSPM.}
\label{fig:i1548394620}
\end{figure*}

\begin{figure*}[t!]
\centering
\includegraphics[width=\linewidth]{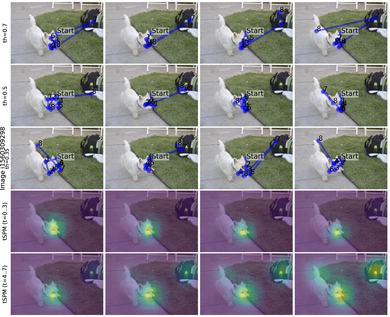}
\caption{Additional results for image \emph{i1560309298}. (First row) Four predicted scanpaths with $th=0.7$. (Second row) Four predicted scanpaths with $th=0.5$. (Third row) Four predicted scanpaths with $th=0.35$. (Fourth and fifth row) Predicted sequence of tSPM.}
\label{fig:i1560309298}
\end{figure*}

\begin{figure*}[t!]
\centering
\includegraphics[width=\linewidth]{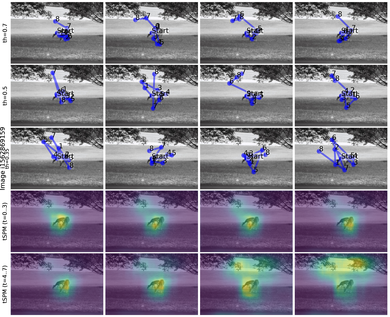}
\caption{Additional results for image \emph{i1562869159}. (First row) Four predicted scanpaths with $th=0.7$. (Second row) Four predicted scanpaths with $th=0.5$. (Third row) Four predicted scanpaths with $th=0.35$. (Fourth and fifth row) Predicted sequence of tSPM.}
\label{fig:i1562869159}
\end{figure*}

\begin{figure*}[t!]
\centering
\includegraphics[width=\linewidth]{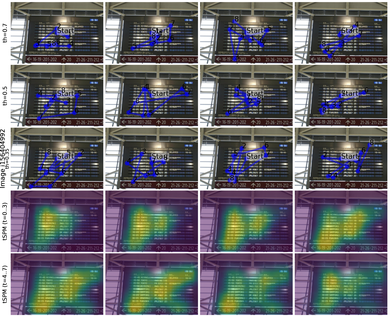}
\caption{Additional results for image \emph{i156404992}. (First row) Four predicted scanpaths with $th=0.7$. (Second row) Four predicted scanpaths with $th=0.5$. (Third row) Four predicted scanpaths with $th=0.35$. (Fourth and fifth row) Predicted sequence of tSPM.}
\label{fig:i156404992}
\end{figure*}

\begin{figure*}[t!]
\centering
\includegraphics[width=\linewidth]{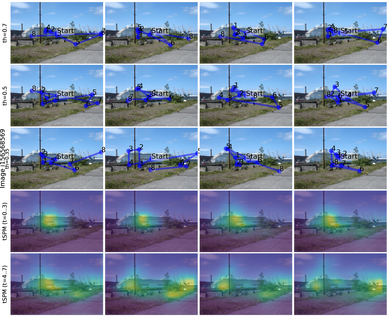}
\caption{Additional results for image \emph{i156568569}. (First row) Four predicted scanpaths with $th=0.7$. (Second row) Four predicted scanpaths with $th=0.5$. (Third row) Four predicted scanpaths with $th=0.35$. (Fourth and fifth row) Predicted sequence of tSPM.}
\label{fig:i156568569}
\end{figure*}

\begin{figure*}[t!]
\centering
\includegraphics[width=\linewidth]{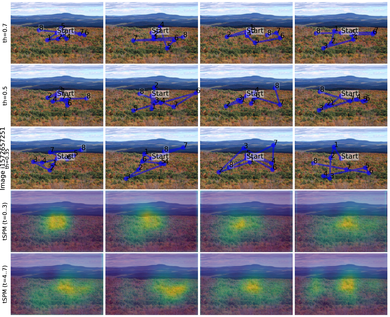}
\caption{Additional results for image \emph{i1572657251}. (First row) Four predicted scanpaths with $th=0.7$. (Second row) Four predicted scanpaths with $th=0.5$. (Third row) Four predicted scanpaths with $th=0.35$. (Fourth and fifth row) Predicted sequence of tSPM.}
\label{fig:i1572657251}
\end{figure*}

\begin{figure*}[t!]
\centering
\includegraphics[width=\linewidth]{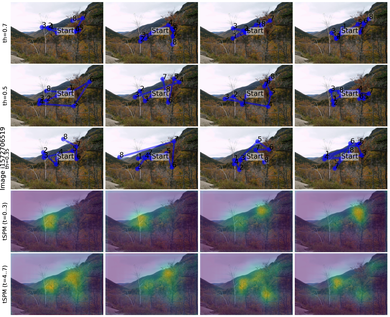}
\caption{Additional results for image \emph{i1572706519}. (First row) Four predicted scanpaths with $th=0.7$. (Second row) Four predicted scanpaths with $th=0.5$. (Third row) Four predicted scanpaths with $th=0.35$. (Fourth and fifth row) Predicted sequence of tSPM.}
\label{fig:i1572706519}
\end{figure*}

\begin{figure*}[t!]
\centering
\includegraphics[width=\linewidth]{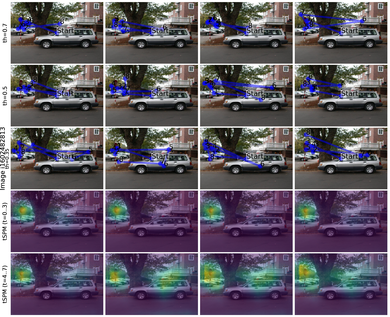}
\caption{Additional results for image \emph{i1602482813}. (First row) Four predicted scanpaths with $th=0.7$. (Second row) Four predicted scanpaths with $th=0.5$. (Third row) Four predicted scanpaths with $th=0.35$. (Fourth and fifth row) Predicted sequence of tSPM.}
\label{fig:i1602482813}
\end{figure*}

\begin{figure*}[t!]
\centering
\includegraphics[width=\linewidth]{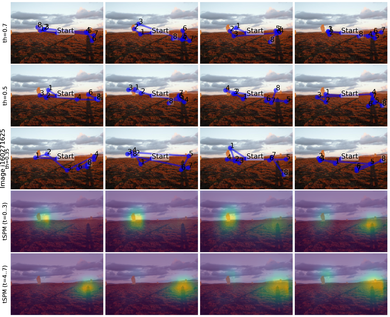}
\caption{Additional results for image \emph{i160271625}. (First row) Four predicted scanpaths with $th=0.7$. (Second row) Four predicted scanpaths with $th=0.5$. (Third row) Four predicted scanpaths with $th=0.35$. (Fourth and fifth row) Predicted sequence of tSPM.}
\label{fig:i160271625}
\end{figure*}

\begin{figure*}[t!]
\centering
\includegraphics[width=\linewidth]{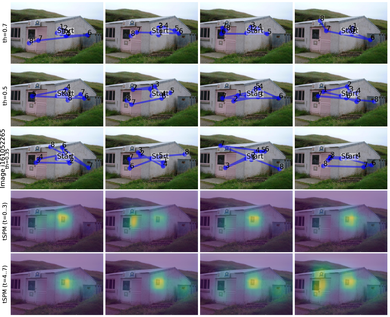}
\caption{Additional results for image \emph{i161052265}. (First row) Four predicted scanpaths with $th=0.7$. (Second row) Four predicted scanpaths with $th=0.5$. (Third row) Four predicted scanpaths with $th=0.35$. (Fourth and fifth row) Predicted sequence of tSPM.}
\label{fig:i161052265}
\end{figure*}

\begin{figure*}[t!]
\centering
\includegraphics[width=\linewidth]{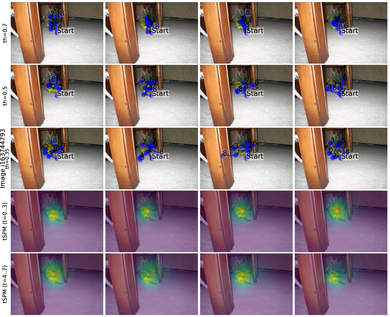}
\caption{Additional results for image \emph{i163744793}. (First row) Four predicted scanpaths with $th=0.7$. (Second row) Four predicted scanpaths with $th=0.5$. (Third row) Four predicted scanpaths with $th=0.35$. (Fourth and fifth row) Predicted sequence of tSPM.}
\label{fig:i163744793}
\end{figure*}

\begin{figure*}[t!]
\centering
\includegraphics[width=\linewidth]{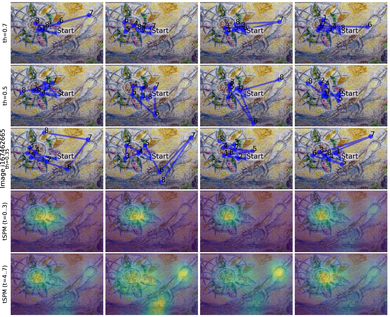}
\caption{Additional results for image \emph{i167462665}. (First row) Four predicted scanpaths with $th=0.7$. (Second row) Four predicted scanpaths with $th=0.5$. (Third row) Four predicted scanpaths with $th=0.35$. (Fourth and fifth row) Predicted sequence of tSPM.}
\label{fig:i167462665}
\end{figure*}

\begin{figure*}[t!]
\centering
\includegraphics[width=\linewidth]{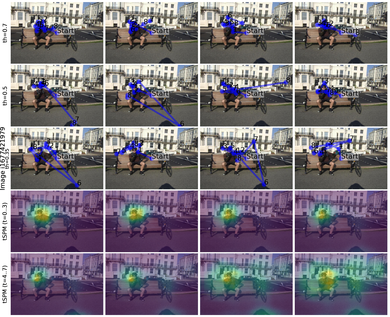}
\caption{Additional results for image \emph{i1677421979}. (First row) Four predicted scanpaths with $th=0.7$. (Second row) Four predicted scanpaths with $th=0.5$. (Third row) Four predicted scanpaths with $th=0.35$. (Fourth and fifth row) Predicted sequence of tSPM.}
\label{fig:i1677421979}
\end{figure*}

\begin{figure*}[t!]
\centering
\includegraphics[width=\linewidth]{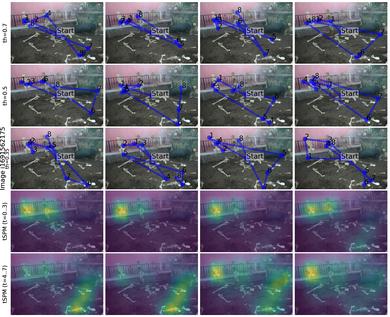}
\caption{Additional results for image \emph{i1691562175}. (First row) Four predicted scanpaths with $th=0.7$. (Second row) Four predicted scanpaths with $th=0.5$. (Third row) Four predicted scanpaths with $th=0.35$. (Fourth and fifth row) Predicted sequence of tSPM.}
\label{fig:i1691562175}
\end{figure*}

\begin{figure*}[t!]
\centering
\includegraphics[width=\linewidth]{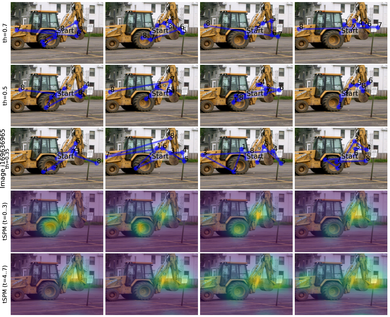}
\caption{Additional results for image \emph{i169636965}. (First row) Four predicted scanpaths with $th=0.7$. (Second row) Four predicted scanpaths with $th=0.5$. (Third row) Four predicted scanpaths with $th=0.35$. (Fourth and fifth row) Predicted sequence of tSPM.}
\label{fig:i169636965}
\end{figure*}

\begin{figure*}[t!]
\centering
\includegraphics[width=\linewidth]{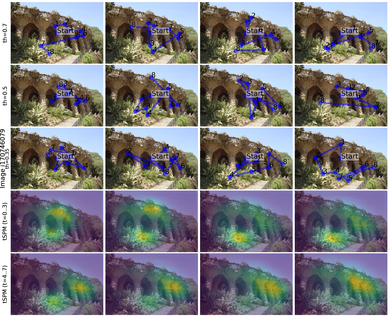}
\caption{Additional results for image \emph{i170746079}. (First row) Four predicted scanpaths with $th=0.7$. (Second row) Four predicted scanpaths with $th=0.5$. (Third row) Four predicted scanpaths with $th=0.35$. (Fourth and fifth row) Predicted sequence of tSPM.}
\label{fig:i170746079}
\end{figure*}

\begin{figure*}[t!]
\centering
\includegraphics[width=\linewidth]{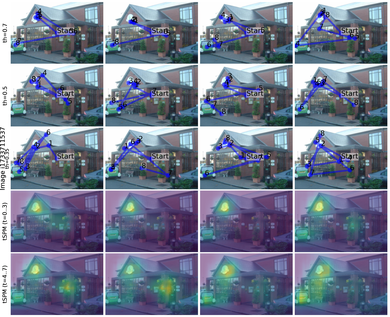}
\caption{Additional results for image \emph{i1733711537}. (First row) Four predicted scanpaths with $th=0.7$. (Second row) Four predicted scanpaths with $th=0.5$. (Third row) Four predicted scanpaths with $th=0.35$. (Fourth and fifth row) Predicted sequence of tSPM.}
\label{fig:i1733711537}
\end{figure*}

\begin{figure*}[t!]
\centering
\includegraphics[width=\linewidth]{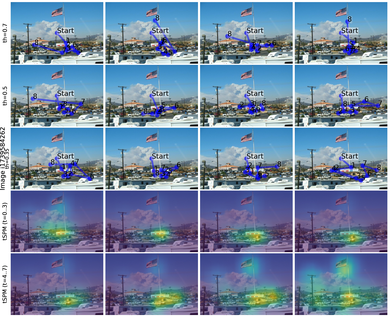}
\caption{Additional results for image \emph{i1739584262}. (First row) Four predicted scanpaths with $th=0.7$. (Second row) Four predicted scanpaths with $th=0.5$. (Third row) Four predicted scanpaths with $th=0.35$. (Fourth and fifth row) Predicted sequence of tSPM.}
\label{fig:i1739584262}
\end{figure*}

\begin{figure*}[t!]
\centering
\includegraphics[width=\linewidth]{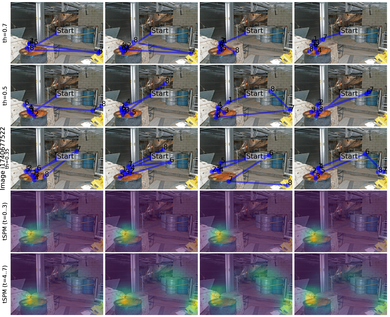}
\caption{Additional results for image \emph{i1740677522}. (First row) Four predicted scanpaths with $th=0.7$. (Second row) Four predicted scanpaths with $th=0.5$. (Third row) Four predicted scanpaths with $th=0.35$. (Fourth and fifth row) Predicted sequence of tSPM.}
\label{fig:i1740677522}
\end{figure*}

\begin{figure*}[t!]
\centering
\includegraphics[width=\linewidth]{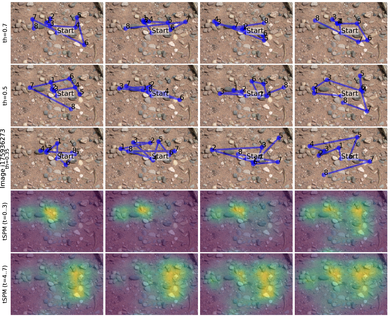}
\caption{Additional results for image \emph{i175936273}. (First row) Four predicted scanpaths with $th=0.7$. (Second row) Four predicted scanpaths with $th=0.5$. (Third row) Four predicted scanpaths with $th=0.35$. (Fourth and fifth row) Predicted sequence of tSPM.}
\label{fig:i175936273}
\end{figure*}

\begin{figure*}[t!]
\centering
\includegraphics[width=\linewidth]{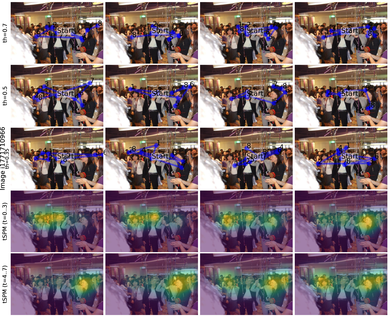}
\caption{Additional results for image \emph{i1771710966}. (First row) Four predicted scanpaths with $th=0.7$. (Second row) Four predicted scanpaths with $th=0.5$. (Third row) Four predicted scanpaths with $th=0.35$. (Fourth and fifth row) Predicted sequence of tSPM.}
\label{fig:i1771710966}
\end{figure*}

\begin{figure*}[t!]
\centering
\includegraphics[width=\linewidth]{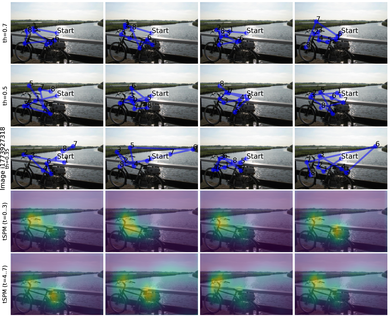}
\caption{Additional results for image \emph{i1773927318}. (First row) Four predicted scanpaths with $th=0.7$. (Second row) Four predicted scanpaths with $th=0.5$. (Third row) Four predicted scanpaths with $th=0.35$. (Fourth and fifth row) Predicted sequence of tSPM.}
\label{fig:i1773927318}
\end{figure*}

\begin{figure*}[t!]
\centering
\includegraphics[width=\linewidth]{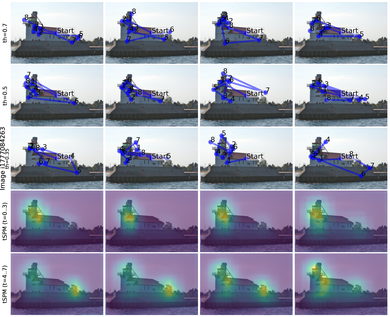}
\caption{Additional results for image \emph{i1777084263}. (First row) Four predicted scanpaths with $th=0.7$. (Second row) Four predicted scanpaths with $th=0.5$. (Third row) Four predicted scanpaths with $th=0.35$. (Fourth and fifth row) Predicted sequence of tSPM.}
\label{fig:i1777084263}
\end{figure*}

\begin{figure*}[t!]
\centering
\includegraphics[width=\linewidth]{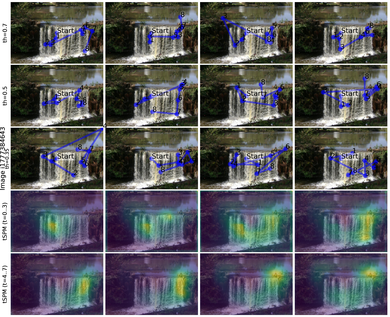}
\caption{Additional results for image \emph{i1777384643}. (First row) Four predicted scanpaths with $th=0.7$. (Second row) Four predicted scanpaths with $th=0.5$. (Third row) Four predicted scanpaths with $th=0.35$. (Fourth and fifth row) Predicted sequence of tSPM.}
\label{fig:i1777384643}
\end{figure*}

\begin{figure*}[t!]
\centering
\includegraphics[width=\linewidth]{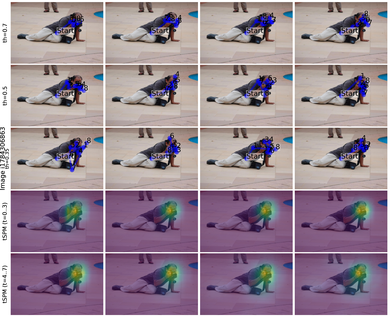}
\caption{Additional results for image \emph{i1784306863}. (First row) Four predicted scanpaths with $th=0.7$. (Second row) Four predicted scanpaths with $th=0.5$. (Third row) Four predicted scanpaths with $th=0.35$. (Fourth and fifth row) Predicted sequence of tSPM.}
\label{fig:i1784306863}
\end{figure*}

\begin{figure*}[t!]
\centering
\includegraphics[width=\linewidth]{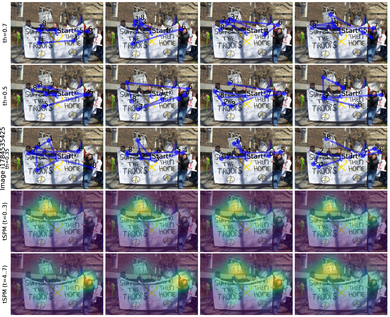}
\caption{Additional results for image \emph{i1784535425}. (First row) Four predicted scanpaths with $th=0.7$. (Second row) Four predicted scanpaths with $th=0.5$. (Third row) Four predicted scanpaths with $th=0.35$. (Fourth and fifth row) Predicted sequence of tSPM.}
\label{fig:i1784535425}
\end{figure*}

\begin{figure*}[t!]
\centering
\includegraphics[width=\linewidth]{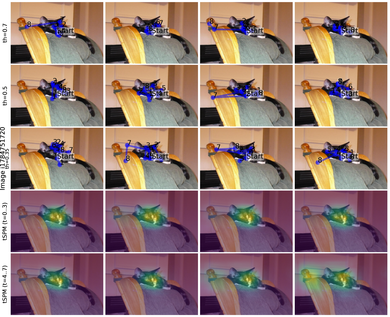}
\caption{Additional results for image \emph{i1784751720}. (First row) Four predicted scanpaths with $th=0.7$. (Second row) Four predicted scanpaths with $th=0.5$. (Third row) Four predicted scanpaths with $th=0.35$. (Fourth and fifth row) Predicted sequence of tSPM.}
\label{fig:i1784751720}
\end{figure*}

\begin{figure*}[t!]
\centering
\includegraphics[width=\linewidth]{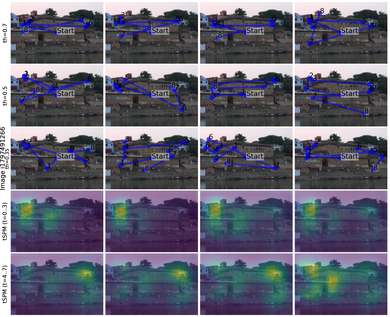}
\caption{Additional results for image \emph{i1797491266}. (First row) Four predicted scanpaths with $th=0.7$. (Second row) Four predicted scanpaths with $th=0.5$. (Third row) Four predicted scanpaths with $th=0.35$. (Fourth and fifth row) Predicted sequence of tSPM.}
\label{fig:i1797491266}
\end{figure*}

\begin{figure*}[t!]
\centering
\includegraphics[width=\linewidth]{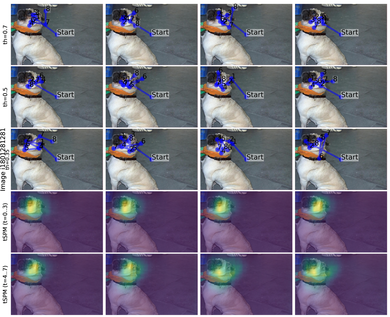}
\caption{Additional results for image \emph{i1801281281}. (First row) Four predicted scanpaths with $th=0.7$. (Second row) Four predicted scanpaths with $th=0.5$. (Third row) Four predicted scanpaths with $th=0.35$. (Fourth and fifth row) Predicted sequence of tSPM.}
\label{fig:i1801281281}
\end{figure*}

\begin{figure*}[t!]
\centering
\includegraphics[width=\linewidth]{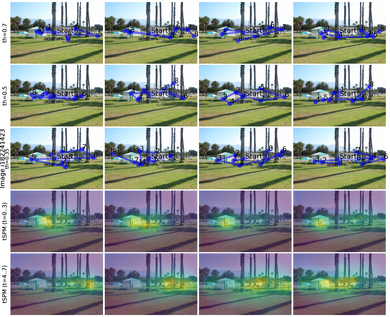}
\caption{Additional results for image \emph{i182241423}. (First row) Four predicted scanpaths with $th=0.7$. (Second row) Four predicted scanpaths with $th=0.5$. (Third row) Four predicted scanpaths with $th=0.35$. (Fourth and fifth row) Predicted sequence of tSPM.}
\label{fig:i182241423}
\end{figure*}

\begin{figure*}[t!]
\centering
\includegraphics[width=\linewidth]{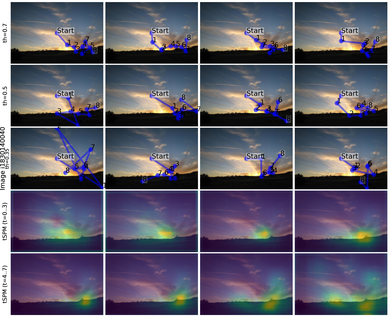}
\caption{Additional results for image \emph{i1830140040}. (First row) Four predicted scanpaths with $th=0.7$. (Second row) Four predicted scanpaths with $th=0.5$. (Third row) Four predicted scanpaths with $th=0.35$. (Fourth and fifth row) Predicted sequence of tSPM.}
\label{fig:i1830140040}
\end{figure*}

\begin{figure*}[t!]
\centering
\includegraphics[width=\linewidth]{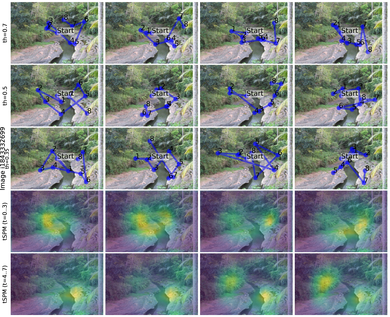}
\caption{Additional results for image \emph{i1843332699}. (First row) Four predicted scanpaths with $th=0.7$. (Second row) Four predicted scanpaths with $th=0.5$. (Third row) Four predicted scanpaths with $th=0.35$. (Fourth and fifth row) Predicted sequence of tSPM.}
\label{fig:i1843332699}
\end{figure*}

\begin{figure*}[t!]
\centering
\includegraphics[width=\linewidth]{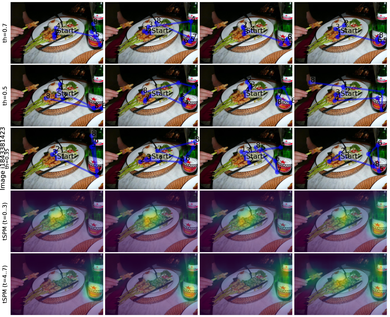}
\caption{Additional results for image \emph{i1843381423}. (First row) Four predicted scanpaths with $th=0.7$. (Second row) Four predicted scanpaths with $th=0.5$. (Third row) Four predicted scanpaths with $th=0.35$. (Fourth and fifth row) Predicted sequence of tSPM.}
\label{fig:i1843381423}
\end{figure*}

\begin{figure*}[t!]
\centering
\includegraphics[width=\linewidth]{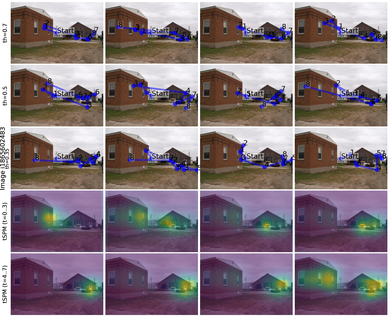}
\caption{Additional results for image \emph{i1865602483}. (First row) Four predicted scanpaths with $th=0.7$. (Second row) Four predicted scanpaths with $th=0.5$. (Third row) Four predicted scanpaths with $th=0.35$. (Fourth and fifth row) Predicted sequence of tSPM.}
\label{fig:i1865602483}
\end{figure*}

\begin{figure*}[t!]
\centering
\includegraphics[width=\linewidth]{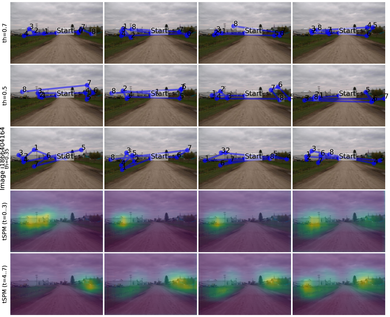}
\caption{Additional results for image \emph{i1866404164}. (First row) Four predicted scanpaths with $th=0.7$. (Second row) Four predicted scanpaths with $th=0.5$. (Third row) Four predicted scanpaths with $th=0.35$. (Fourth and fifth row) Predicted sequence of tSPM.}
\label{fig:i1866404164}
\end{figure*}

\begin{figure*}[t!]
\centering
\includegraphics[width=\linewidth]{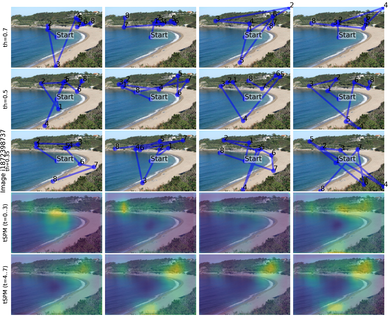}
\caption{Additional results for image \emph{i1872398737}. (First row) Four predicted scanpaths with $th=0.7$. (Second row) Four predicted scanpaths with $th=0.5$. (Third row) Four predicted scanpaths with $th=0.35$. (Fourth and fifth row) Predicted sequence of tSPM.}
\label{fig:i1872398737}
\end{figure*}

\begin{figure*}[t!]
\centering
\includegraphics[width=\linewidth]{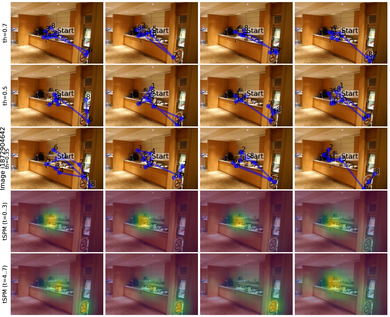}
\caption{Additional results for image \emph{i1872904642}. (First row) Four predicted scanpaths with $th=0.7$. (Second row) Four predicted scanpaths with $th=0.5$. (Third row) Four predicted scanpaths with $th=0.35$. (Fourth and fifth row) Predicted sequence of tSPM.}
\label{fig:i1872904642}
\end{figure*}

\begin{figure*}[t!]
\centering
\includegraphics[width=\linewidth]{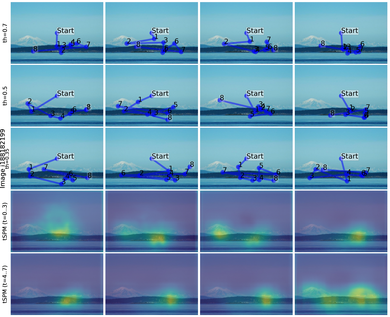}
\caption{Additional results for image \emph{i188182199}. (First row) Four predicted scanpaths with $th=0.7$. (Second row) Four predicted scanpaths with $th=0.5$. (Third row) Four predicted scanpaths with $th=0.35$. (Fourth and fifth row) Predicted sequence of tSPM.}
\label{fig:i188182199}
\end{figure*}

\begin{figure*}[t!]
\centering
\includegraphics[width=\linewidth]{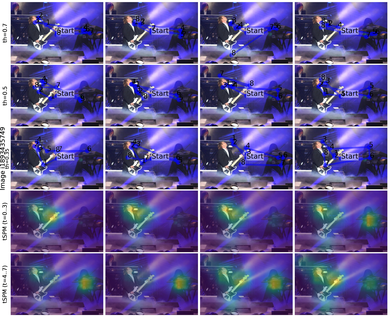}
\caption{Additional results for image \emph{i1893435749}. (First row) Four predicted scanpaths with $th=0.7$. (Second row) Four predicted scanpaths with $th=0.5$. (Third row) Four predicted scanpaths with $th=0.35$. (Fourth and fifth row) Predicted sequence of tSPM.}
\label{fig:i1893435749}
\end{figure*}

\begin{figure*}[t!]
\centering
\includegraphics[width=\linewidth]{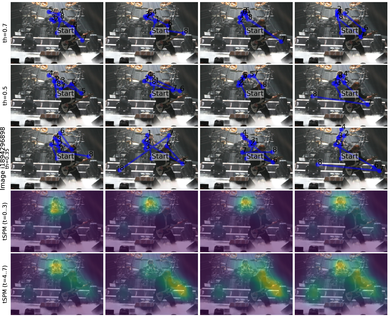}
\caption{Additional results for image \emph{i1894296898}. (First row) Four predicted scanpaths with $th=0.7$. (Second row) Four predicted scanpaths with $th=0.5$. (Third row) Four predicted scanpaths with $th=0.35$. (Fourth and fifth row) Predicted sequence of tSPM.}
\label{fig:i1894296898}
\end{figure*}

\begin{figure*}[t!]
\centering
\includegraphics[width=\linewidth]{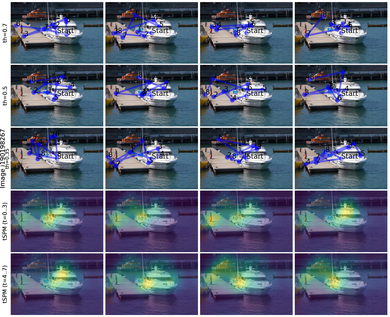}
\caption{Additional results for image \emph{i190198267}. (First row) Four predicted scanpaths with $th=0.7$. (Second row) Four predicted scanpaths with $th=0.5$. (Third row) Four predicted scanpaths with $th=0.35$. (Fourth and fifth row) Predicted sequence of tSPM.}
\label{fig:i190198267}
\end{figure*}

\begin{figure*}[t!]
\centering
\includegraphics[width=\linewidth]{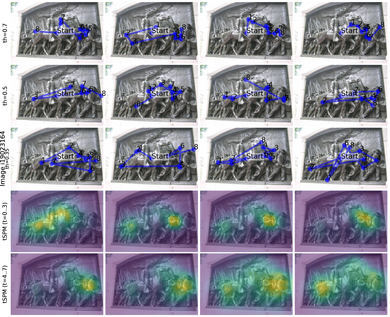}
\caption{Additional results for image \emph{i19023164}. (First row) Four predicted scanpaths with $th=0.7$. (Second row) Four predicted scanpaths with $th=0.5$. (Third row) Four predicted scanpaths with $th=0.35$. (Fourth and fifth row) Predicted sequence of tSPM.}
\label{fig:i19023164}
\end{figure*}

\begin{figure*}[t!]
\centering
\includegraphics[width=\linewidth]{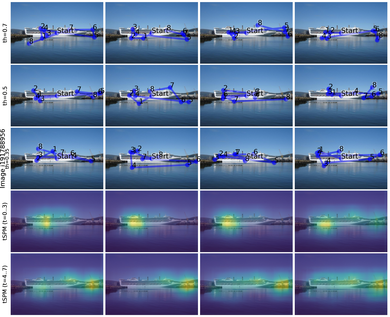}
\caption{Additional results for image \emph{i191788956}. (First row) Four predicted scanpaths with $th=0.7$. (Second row) Four predicted scanpaths with $th=0.5$. (Third row) Four predicted scanpaths with $th=0.35$. (Fourth and fifth row) Predicted sequence of tSPM.}
\label{fig:i191788956}
\end{figure*}

\begin{figure*}[t!]
\centering
\includegraphics[width=\linewidth]{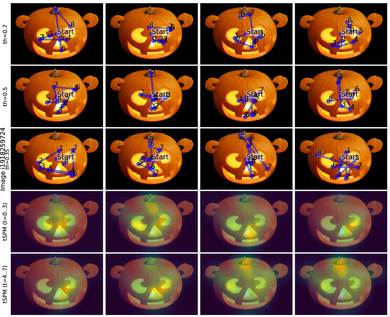}
\caption{Additional results for image \emph{i1918259724}. (First row) Four predicted scanpaths with $th=0.7$. (Second row) Four predicted scanpaths with $th=0.5$. (Third row) Four predicted scanpaths with $th=0.35$. (Fourth and fifth row) Predicted sequence of tSPM.}
\label{fig:i1918259724}
\end{figure*}

\begin{figure*}[t!]
\centering
\includegraphics[width=\linewidth]{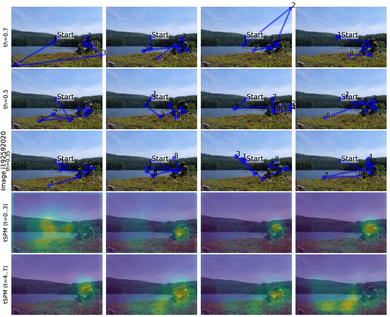}
\caption{Additional results for image \emph{i192592020}. (First row) Four predicted scanpaths with $th=0.7$. (Second row) Four predicted scanpaths with $th=0.5$. (Third row) Four predicted scanpaths with $th=0.35$. (Fourth and fifth row) Predicted sequence of tSPM.}
\label{fig:i192592020}
\end{figure*}

\begin{figure*}[t!]
\centering
\includegraphics[width=\linewidth]{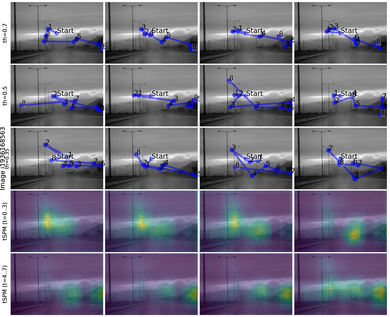}
\caption{Additional results for image \emph{i1936168563}. (First row) Four predicted scanpaths with $th=0.7$. (Second row) Four predicted scanpaths with $th=0.5$. (Third row) Four predicted scanpaths with $th=0.35$. (Fourth and fifth row) Predicted sequence of tSPM.}
\label{fig:i1936168563}
\end{figure*}

\begin{figure*}[t!]
\centering
\includegraphics[width=\linewidth]{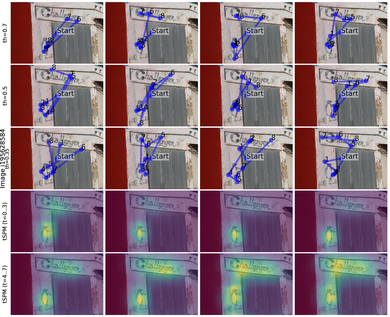}
\caption{Additional results for image \emph{i195628584}. (First row) Four predicted scanpaths with $th=0.7$. (Second row) Four predicted scanpaths with $th=0.5$. (Third row) Four predicted scanpaths with $th=0.35$. (Fourth and fifth row) Predicted sequence of tSPM.}
\label{fig:i195628584}
\end{figure*}

\begin{figure*}[t!]
\centering
\includegraphics[width=\linewidth]{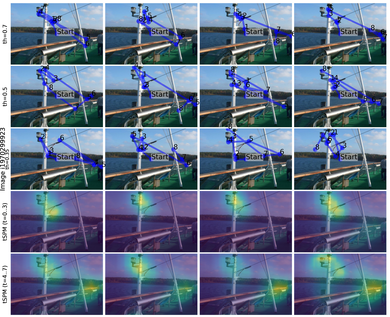}
\caption{Additional results for image \emph{i1970299923}. (First row) Four predicted scanpaths with $th=0.7$. (Second row) Four predicted scanpaths with $th=0.5$. (Third row) Four predicted scanpaths with $th=0.35$. (Fourth and fifth row) Predicted sequence of tSPM.}
\label{fig:i1970299923}
\end{figure*}

\begin{figure*}[t!]
\centering
\includegraphics[width=\linewidth]{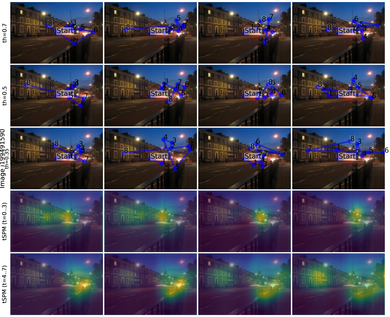}
\caption{Additional results for image \emph{i199491590}. (First row) Four predicted scanpaths with $th=0.7$. (Second row) Four predicted scanpaths with $th=0.5$. (Third row) Four predicted scanpaths with $th=0.35$. (Fourth and fifth row) Predicted sequence of tSPM.}
\label{fig:i199491590}
\end{figure*}

\begin{figure*}[t!]
\centering
\includegraphics[width=\linewidth]{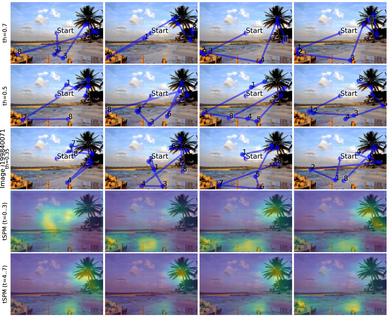}
\caption{Additional results for image \emph{i199840071}. (First row) Four predicted scanpaths with $th=0.7$. (Second row) Four predicted scanpaths with $th=0.5$. (Third row) Four predicted scanpaths with $th=0.35$. (Fourth and fifth row) Predicted sequence of tSPM.}
\label{fig:i199840071}
\end{figure*}

\begin{figure*}[t!]
\centering
\includegraphics[width=\linewidth]{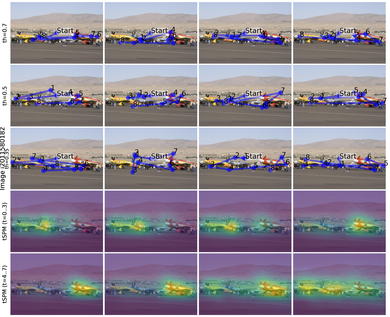}
\caption{Additional results for image \emph{i2011580182}. (First row) Four predicted scanpaths with $th=0.7$. (Second row) Four predicted scanpaths with $th=0.5$. (Third row) Four predicted scanpaths with $th=0.35$. (Fourth and fifth row) Predicted sequence of tSPM.}
\label{fig:i2011580182}
\end{figure*}

\begin{figure*}[t!]
\centering
\includegraphics[width=\linewidth]{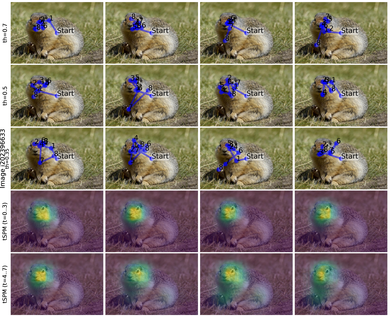}
\caption{Additional results for image \emph{i202396633}. (First row) Four predicted scanpaths with $th=0.7$. (Second row) Four predicted scanpaths with $th=0.5$. (Third row) Four predicted scanpaths with $th=0.35$. (Fourth and fifth row) Predicted sequence of tSPM.}
\label{fig:i202396633}
\end{figure*}

\begin{figure*}[t!]
\centering
\includegraphics[width=\linewidth]{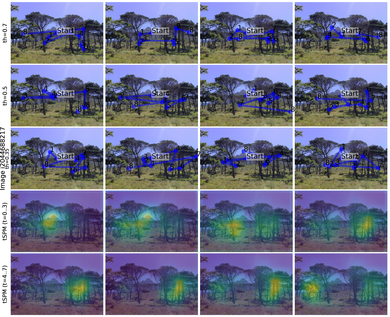}
\caption{Additional results for image \emph{i2044688217}. (First row) Four predicted scanpaths with $th=0.7$. (Second row) Four predicted scanpaths with $th=0.5$. (Third row) Four predicted scanpaths with $th=0.35$. (Fourth and fifth row) Predicted sequence of tSPM.}
\label{fig:i2044688217}
\end{figure*}

\begin{figure*}[t!]
\centering
\includegraphics[width=\linewidth]{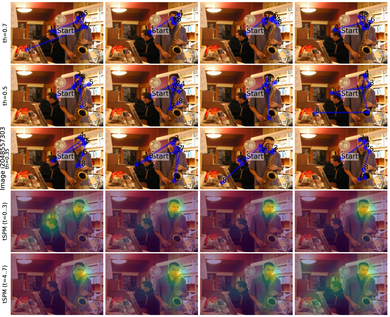}
\caption{Additional results for image \emph{i2048557303}. (First row) Four predicted scanpaths with $th=0.7$. (Second row) Four predicted scanpaths with $th=0.5$. (Third row) Four predicted scanpaths with $th=0.35$. (Fourth and fifth row) Predicted sequence of tSPM.}
\label{fig:i2048557303}
\end{figure*}

\begin{figure*}[t!]
\centering
\includegraphics[width=\linewidth]{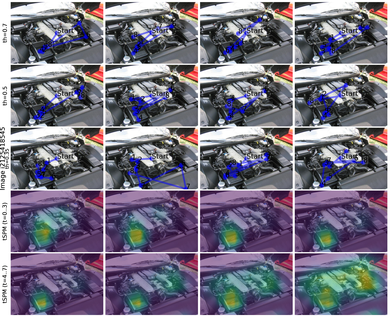}
\caption{Additional results for image \emph{i2125418545}. (First row) Four predicted scanpaths with $th=0.7$. (Second row) Four predicted scanpaths with $th=0.5$. (Third row) Four predicted scanpaths with $th=0.35$. (Fourth and fifth row) Predicted sequence of tSPM.}
\label{fig:i2125418545}
\end{figure*}

\begin{figure*}[t!]
\centering
\includegraphics[width=\linewidth]{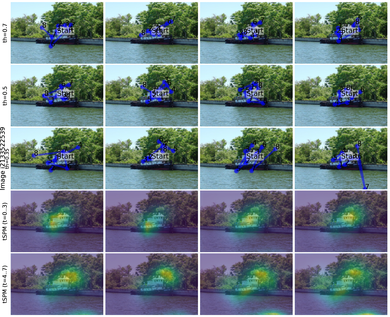}
\caption{Additional results for image \emph{i2133522539}. (First row) Four predicted scanpaths with $th=0.7$. (Second row) Four predicted scanpaths with $th=0.5$. (Third row) Four predicted scanpaths with $th=0.35$. (Fourth and fifth row) Predicted sequence of tSPM.}
\label{fig:i2133522539}
\end{figure*}

\begin{figure*}[t!]
\centering
\includegraphics[width=\linewidth]{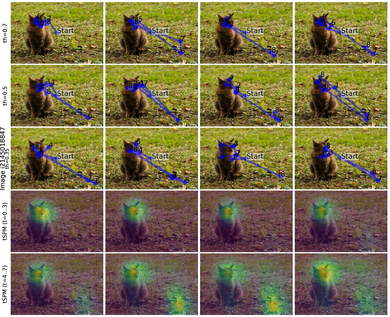}
\caption{Additional results for image \emph{i2145018847}. (First row) Four predicted scanpaths with $th=0.7$. (Second row) Four predicted scanpaths with $th=0.5$. (Third row) Four predicted scanpaths with $th=0.35$. (Fourth and fifth row) Predicted sequence of tSPM.}
\label{fig:i2145018847}
\end{figure*}

\begin{figure*}[t!]
\centering
\includegraphics[width=\linewidth]{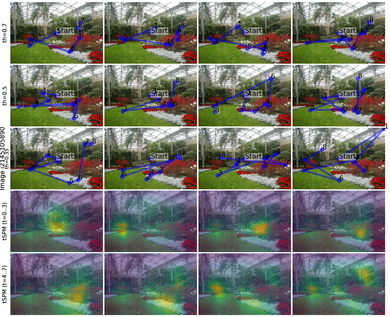}
\caption{Additional results for image \emph{i2145105890}. (First row) Four predicted scanpaths with $th=0.7$. (Second row) Four predicted scanpaths with $th=0.5$. (Third row) Four predicted scanpaths with $th=0.35$. (Fourth and fifth row) Predicted sequence of tSPM.}
\label{fig:i2145105890}
\end{figure*}

\begin{figure*}[t!]
\centering
\includegraphics[width=\linewidth]{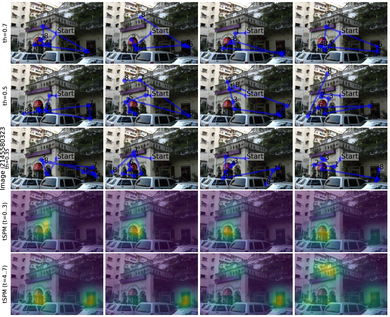}
\caption{Additional results for image \emph{i2145580323}. (First row) Four predicted scanpaths with $th=0.7$. (Second row) Four predicted scanpaths with $th=0.5$. (Third row) Four predicted scanpaths with $th=0.35$. (Fourth and fifth row) Predicted sequence of tSPM.}
\label{fig:i2145580323}
\end{figure*}

\begin{figure*}[t!]
\centering
\includegraphics[width=\linewidth]{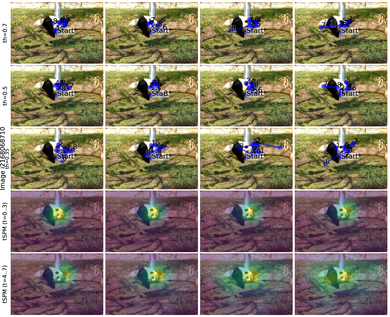}
\caption{Additional results for image \emph{i2168068710}. (First row) Four predicted scanpaths with $th=0.7$. (Second row) Four predicted scanpaths with $th=0.5$. (Third row) Four predicted scanpaths with $th=0.35$. (Fourth and fifth row) Predicted sequence of tSPM.}
\label{fig:i2168068710}
\end{figure*}

\begin{figure*}[t!]
\centering
\includegraphics[width=\linewidth]{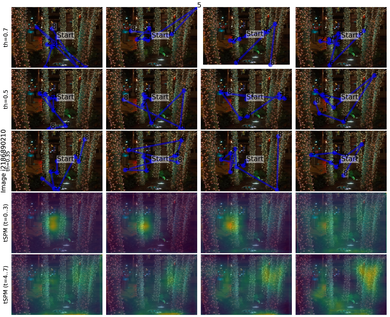}
\caption{Additional results for image \emph{i2186890210}. (First row) Four predicted scanpaths with $th=0.7$. (Second row) Four predicted scanpaths with $th=0.5$. (Third row) Four predicted scanpaths with $th=0.35$. (Fourth and fifth row) Predicted sequence of tSPM.}
\label{fig:i2186890210}
\end{figure*}

\begin{figure*}[t!]
\centering
\includegraphics[width=\linewidth]{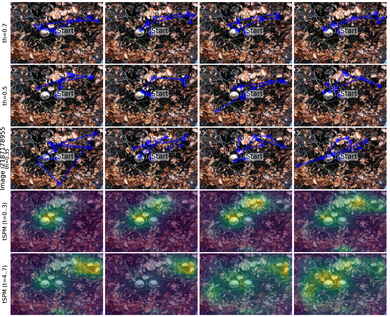}
\caption{Additional results for image \emph{i2187178955}. (First row) Four predicted scanpaths with $th=0.7$. (Second row) Four predicted scanpaths with $th=0.5$. (Third row) Four predicted scanpaths with $th=0.35$. (Fourth and fifth row) Predicted sequence of tSPM.}
\label{fig:i2187178955}
\end{figure*}

\begin{figure*}[t!]
\centering
\includegraphics[width=\linewidth]{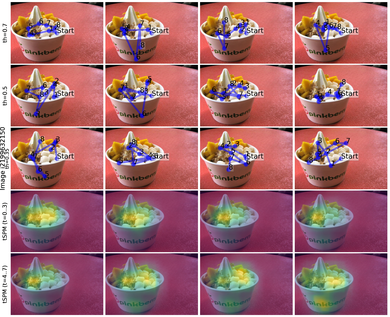}
\caption{Additional results for image \emph{i2199632150}. (First row) Four predicted scanpaths with $th=0.7$. (Second row) Four predicted scanpaths with $th=0.5$. (Third row) Four predicted scanpaths with $th=0.35$. (Fourth and fifth row) Predicted sequence of tSPM.}
\label{fig:i2199632150}
\end{figure*}

\begin{figure*}[t!]
\centering
\includegraphics[width=\linewidth]{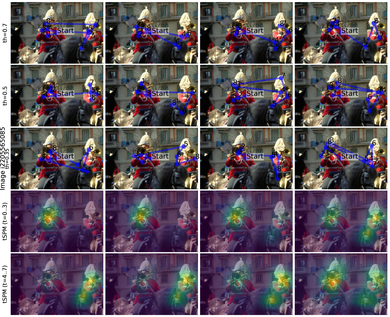}
\caption{Additional results for image \emph{i2205565085}. (First row) Four predicted scanpaths with $th=0.7$. (Second row) Four predicted scanpaths with $th=0.5$. (Third row) Four predicted scanpaths with $th=0.35$. (Fourth and fifth row) Predicted sequence of tSPM.}
\label{fig:i2205565085}
\end{figure*}

\begin{figure*}[t!]
\centering
\includegraphics[width=\linewidth]{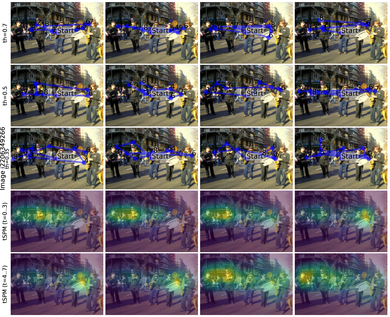}
\caption{Additional results for image \emph{i2206349266}. (First row) Four predicted scanpaths with $th=0.7$. (Second row) Four predicted scanpaths with $th=0.5$. (Third row) Four predicted scanpaths with $th=0.35$. (Fourth and fifth row) Predicted sequence of tSPM.}
\label{fig:i2206349266}
\end{figure*}

\begin{figure*}[t!]
\centering
\includegraphics[width=\linewidth]{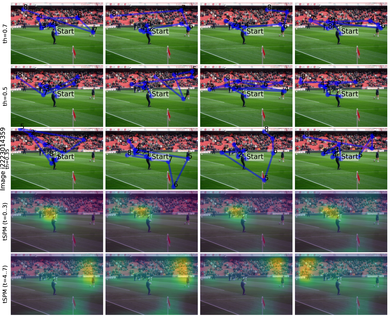}
\caption{Additional results for image \emph{i2223014359}. (First row) Four predicted scanpaths with $th=0.7$. (Second row) Four predicted scanpaths with $th=0.5$. (Third row) Four predicted scanpaths with $th=0.35$. (Fourth and fifth row) Predicted sequence of tSPM.}
\label{fig:i2223014359}
\end{figure*}

\begin{figure*}[t!]
\centering
\includegraphics[width=\linewidth]{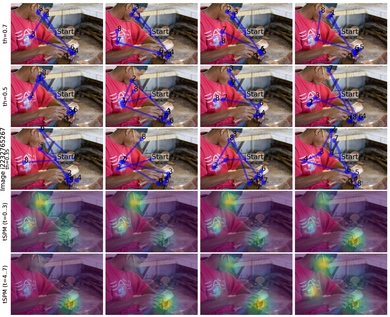}
\caption{Additional results for image \emph{i2232765267}. (First row) Four predicted scanpaths with $th=0.7$. (Second row) Four predicted scanpaths with $th=0.5$. (Third row) Four predicted scanpaths with $th=0.35$. (Fourth and fifth row) Predicted sequence of tSPM.}
\label{fig:i2232765267}
\end{figure*}

\begin{figure*}[t!]
\centering
\includegraphics[width=\linewidth]{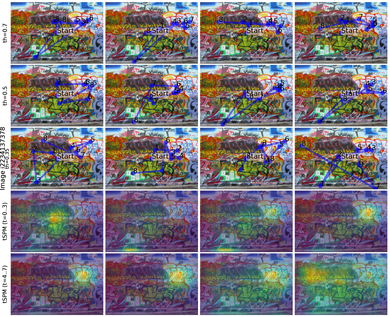}
\caption{Additional results for image \emph{i2234137378}. (First row) Four predicted scanpaths with $th=0.7$. (Second row) Four predicted scanpaths with $th=0.5$. (Third row) Four predicted scanpaths with $th=0.35$. (Fourth and fifth row) Predicted sequence of tSPM.}
\label{fig:i2234137378}
\end{figure*}

\begin{figure*}[t!]
\centering
\includegraphics[width=\linewidth]{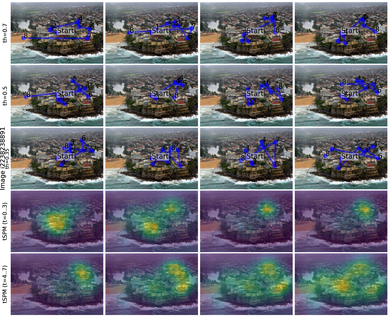}
\caption{Additional results for image \emph{i2238238891}. (First row) Four predicted scanpaths with $th=0.7$. (Second row) Four predicted scanpaths with $th=0.5$. (Third row) Four predicted scanpaths with $th=0.35$. (Fourth and fifth row) Predicted sequence of tSPM.}
\label{fig:i2238238891}
\end{figure*}

\begin{figure*}[t!]
\centering
\includegraphics[width=\linewidth]{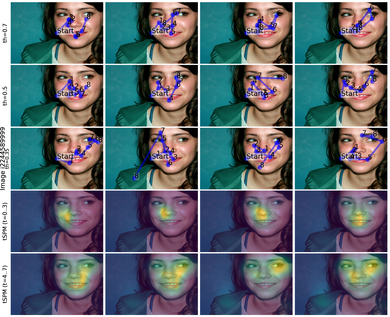}
\caption{Additional results for image \emph{i2244589999}. (First row) Four predicted scanpaths with $th=0.7$. (Second row) Four predicted scanpaths with $th=0.5$. (Third row) Four predicted scanpaths with $th=0.35$. (Fourth and fifth row) Predicted sequence of tSPM.}
\label{fig:i2244589999}
\end{figure*}

\begin{figure*}[t!]
\centering
\includegraphics[width=\linewidth]{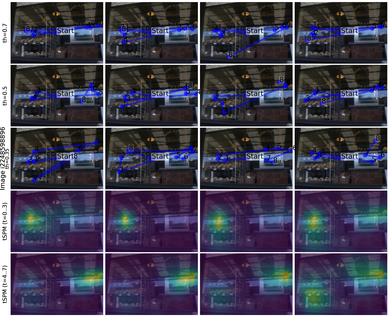}
\caption{Additional results for image \emph{i2248598896}. (First row) Four predicted scanpaths with $th=0.7$. (Second row) Four predicted scanpaths with $th=0.5$. (Third row) Four predicted scanpaths with $th=0.35$. (Fourth and fifth row) Predicted sequence of tSPM.}
\label{fig:i2248598896}
\end{figure*}

\begin{figure*}[t!]
\centering
\includegraphics[width=\linewidth]{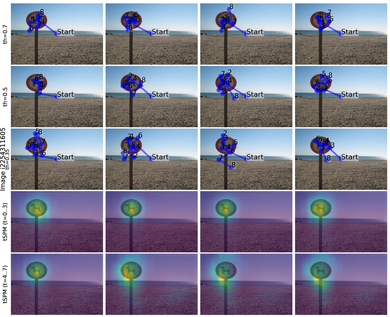}
\caption{Additional results for image \emph{i2254311605}. (First row) Four predicted scanpaths with $th=0.7$. (Second row) Four predicted scanpaths with $th=0.5$. (Third row) Four predicted scanpaths with $th=0.35$. (Fourth and fifth row) Predicted sequence of tSPM.}
\label{fig:i2254311605}
\end{figure*}

\begin{figure*}[t!]
\centering
\includegraphics[width=\linewidth]{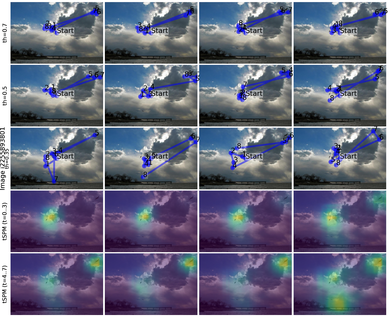}
\caption{Additional results for image \emph{i2255893801}. (First row) Four predicted scanpaths with $th=0.7$. (Second row) Four predicted scanpaths with $th=0.5$. (Third row) Four predicted scanpaths with $th=0.35$. (Fourth and fifth row) Predicted sequence of tSPM.}
\label{fig:i2255893801}
\end{figure*}

\begin{figure*}[t!]
\centering
\includegraphics[width=\linewidth]{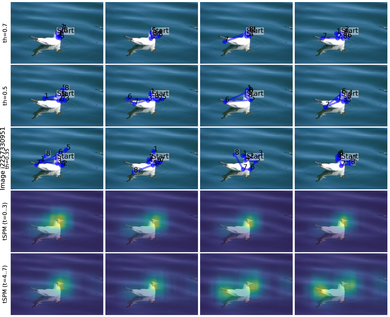}
\caption{Additional results for image \emph{i2257330951}. (First row) Four predicted scanpaths with $th=0.7$. (Second row) Four predicted scanpaths with $th=0.5$. (Third row) Four predicted scanpaths with $th=0.35$. (Fourth and fifth row) Predicted sequence of tSPM.}
\label{fig:i2257330951}
\end{figure*}

\begin{figure*}[t!]
\centering
\includegraphics[width=\linewidth]{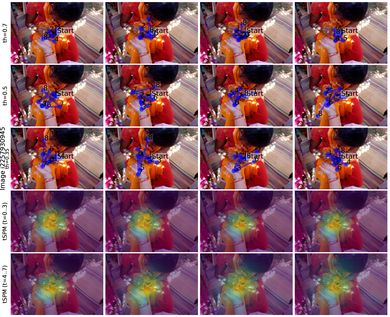}
\caption{Additional results for image \emph{i2257930945}. (First row) Four predicted scanpaths with $th=0.7$. (Second row) Four predicted scanpaths with $th=0.5$. (Third row) Four predicted scanpaths with $th=0.35$. (Fourth and fifth row) Predicted sequence of tSPM.}
\label{fig:i2257930945}
\end{figure*}

\begin{figure*}[t!]
\centering
\includegraphics[width=\linewidth]{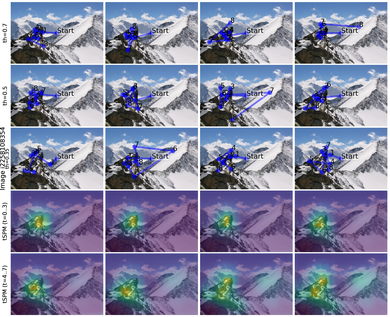}
\caption{Additional results for image \emph{i2258108354}. (First row) Four predicted scanpaths with $th=0.7$. (Second row) Four predicted scanpaths with $th=0.5$. (Third row) Four predicted scanpaths with $th=0.35$. (Fourth and fifth row) Predicted sequence of tSPM.}
\label{fig:i2258108354}
\end{figure*}

\begin{figure*}[t!]
\centering
\includegraphics[width=\linewidth]{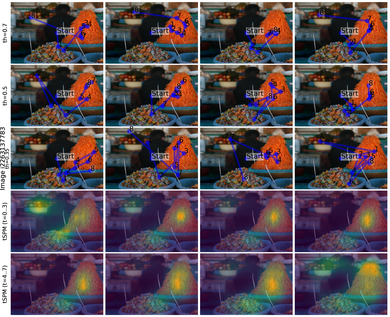}
\caption{Additional results for image \emph{i2263137783}. (First row) Four predicted scanpaths with $th=0.7$. (Second row) Four predicted scanpaths with $th=0.5$. (Third row) Four predicted scanpaths with $th=0.35$. (Fourth and fifth row) Predicted sequence of tSPM.}
\label{fig:i2263137783}
\end{figure*}

\begin{figure*}[t!]
\centering
\includegraphics[width=\linewidth]{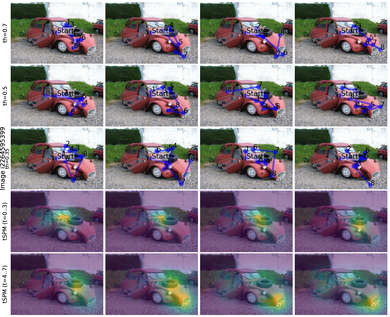}
\caption{Additional results for image \emph{i2264595399}. (First row) Four predicted scanpaths with $th=0.7$. (Second row) Four predicted scanpaths with $th=0.5$. (Third row) Four predicted scanpaths with $th=0.35$. (Fourth and fifth row) Predicted sequence of tSPM.}
\label{fig:i2264595399}
\end{figure*}

\begin{figure*}[t!]
\centering
\includegraphics[width=\linewidth]{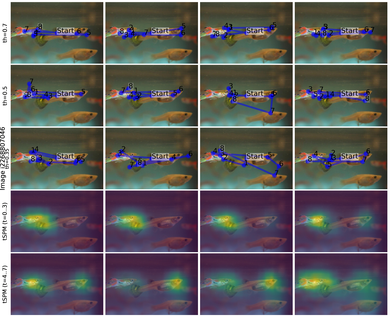}
\caption{Additional results for image \emph{i2268807046}. (First row) Four predicted scanpaths with $th=0.7$. (Second row) Four predicted scanpaths with $th=0.5$. (Third row) Four predicted scanpaths with $th=0.35$. (Fourth and fifth row) Predicted sequence of tSPM.}
\label{fig:i2268807046}
\end{figure*}

\begin{figure*}[t!]
\centering
\includegraphics[width=\linewidth]{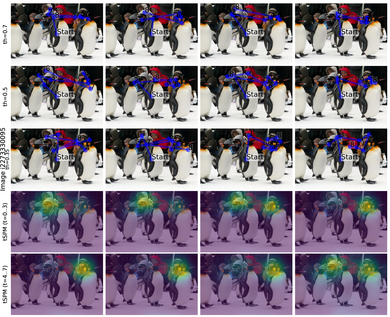}
\caption{Additional results for image \emph{i2273330095}. (First row) Four predicted scanpaths with $th=0.7$. (Second row) Four predicted scanpaths with $th=0.5$. (Third row) Four predicted scanpaths with $th=0.35$. (Fourth and fifth row) Predicted sequence of tSPM.}
\label{fig:i2273330095}
\end{figure*}

\begin{figure*}[t!]
\centering
\includegraphics[width=\linewidth]{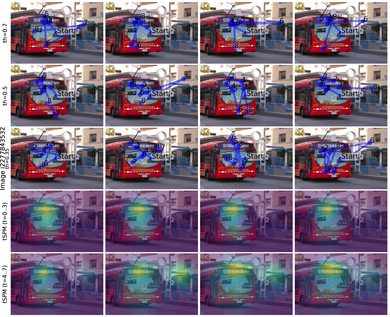}
\caption{Additional results for image \emph{i2277843532}. (First row) Four predicted scanpaths with $th=0.7$. (Second row) Four predicted scanpaths with $th=0.5$. (Third row) Four predicted scanpaths with $th=0.35$. (Fourth and fifth row) Predicted sequence of tSPM.}
\label{fig:i2277843532}
\end{figure*}

\begin{figure*}[t!]
\centering
\includegraphics[width=\linewidth]{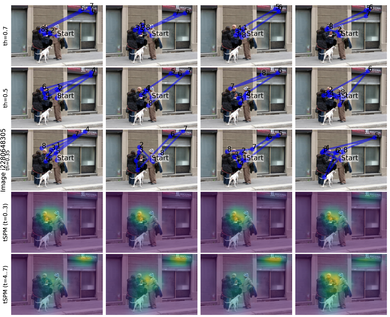}
\caption{Additional results for image \emph{i2280648305}. (First row) Four predicted scanpaths with $th=0.7$. (Second row) Four predicted scanpaths with $th=0.5$. (Third row) Four predicted scanpaths with $th=0.35$. (Fourth and fifth row) Predicted sequence of tSPM.}
\label{fig:i2280648305}
\end{figure*}

\begin{figure*}[t!]
\centering
\includegraphics[width=\linewidth]{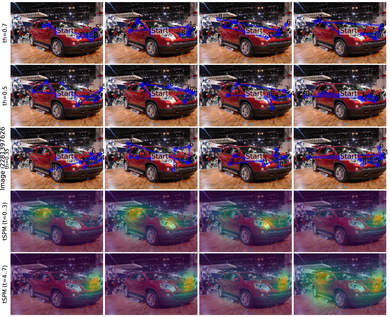}
\caption{Additional results for image \emph{i2281397626}. (First row) Four predicted scanpaths with $th=0.7$. (Second row) Four predicted scanpaths with $th=0.5$. (Third row) Four predicted scanpaths with $th=0.35$. (Fourth and fifth row) Predicted sequence of tSPM.}
\label{fig:i2281397626}
\end{figure*}

\begin{figure*}[t!]
\centering
\includegraphics[width=\linewidth]{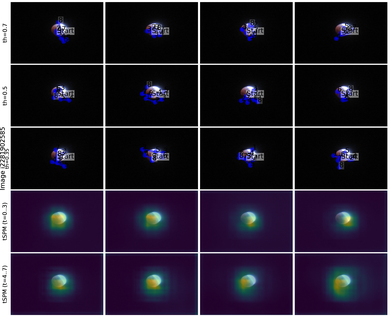}
\caption{Additional results for image \emph{i2281902585}. (First row) Four predicted scanpaths with $th=0.7$. (Second row) Four predicted scanpaths with $th=0.5$. (Third row) Four predicted scanpaths with $th=0.35$. (Fourth and fifth row) Predicted sequence of tSPM.}
\label{fig:i2281902585}
\end{figure*}

\begin{figure*}[t!]
\centering
\includegraphics[width=\linewidth]{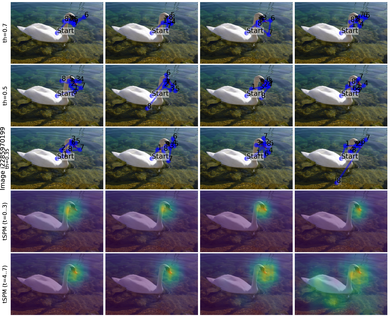}
\caption{Additional results for image \emph{i2285970199}. (First row) Four predicted scanpaths with $th=0.7$. (Second row) Four predicted scanpaths with $th=0.5$. (Third row) Four predicted scanpaths with $th=0.35$. (Fourth and fifth row) Predicted sequence of tSPM.}
\label{fig:i2285970199}
\end{figure*}

\begin{figure*}[t!]
\centering
\includegraphics[width=\linewidth]{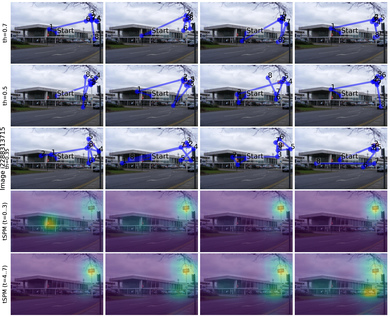}
\caption{Additional results for image \emph{i2288313715}. (First row) Four predicted scanpaths with $th=0.7$. (Second row) Four predicted scanpaths with $th=0.5$. (Third row) Four predicted scanpaths with $th=0.35$. (Fourth and fifth row) Predicted sequence of tSPM.}
\label{fig:i2288313715}
\end{figure*}

\begin{figure*}[t!]
\centering
\includegraphics[width=\linewidth]{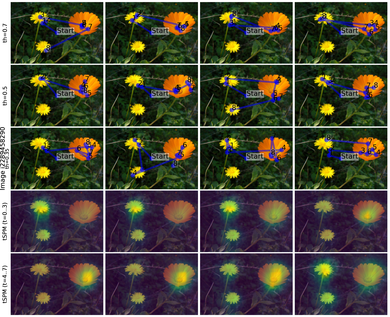}
\caption{Additional results for image \emph{i2289458290}. (First row) Four predicted scanpaths with $th=0.7$. (Second row) Four predicted scanpaths with $th=0.5$. (Third row) Four predicted scanpaths with $th=0.35$. (Fourth and fifth row) Predicted sequence of tSPM.}
\label{fig:i2289458290}
\end{figure*}

\begin{figure*}[t!]
\centering
\includegraphics[width=\linewidth]{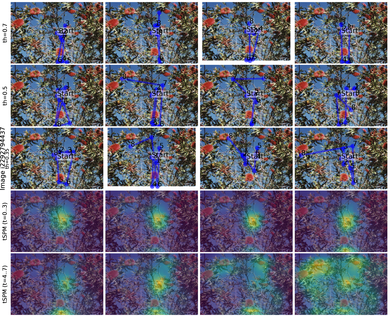}
\caption{Additional results for image \emph{i2292794437}. (First row) Four predicted scanpaths with $th=0.7$. (Second row) Four predicted scanpaths with $th=0.5$. (Third row) Four predicted scanpaths with $th=0.35$. (Fourth and fifth row) Predicted sequence of tSPM.}
\label{fig:i2292794437}
\end{figure*}

\begin{figure*}[t!]
\centering
\includegraphics[width=\linewidth]{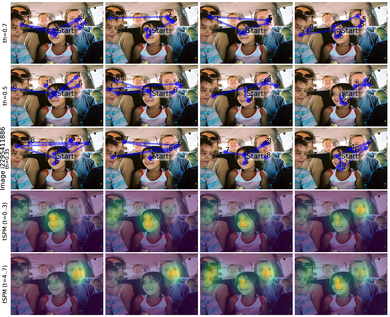}
\caption{Additional results for image \emph{i2295411886}. (First row) Four predicted scanpaths with $th=0.7$. (Second row) Four predicted scanpaths with $th=0.5$. (Third row) Four predicted scanpaths with $th=0.35$. (Fourth and fifth row) Predicted sequence of tSPM.}
\label{fig:i2295411886}
\end{figure*}

\begin{figure*}[t!]
\centering
\includegraphics[width=\linewidth]{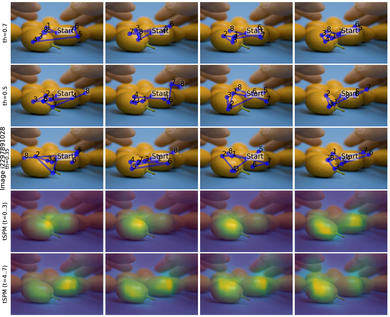}
\caption{Additional results for image \emph{i2297891028}. (First row) Four predicted scanpaths with $th=0.7$. (Second row) Four predicted scanpaths with $th=0.5$. (Third row) Four predicted scanpaths with $th=0.35$. (Fourth and fifth row) Predicted sequence of tSPM.}
\label{fig:i2297891028}
\end{figure*}

\begin{figure*}[t!]
\centering
\includegraphics[width=\linewidth]{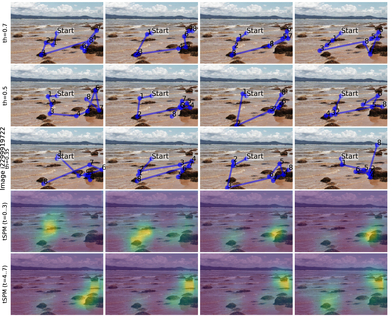}
\caption{Additional results for image \emph{i2299919722}. (First row) Four predicted scanpaths with $th=0.7$. (Second row) Four predicted scanpaths with $th=0.5$. (Third row) Four predicted scanpaths with $th=0.35$. (Fourth and fifth row) Predicted sequence of tSPM.}
\label{fig:i2299919722}
\end{figure*}

\begin{figure*}[t!]
\centering
\includegraphics[width=\linewidth]{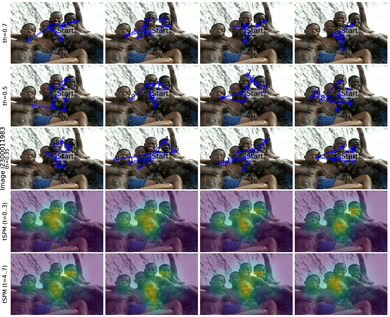}
\caption{Additional results for image \emph{i2300011983}. (First row) Four predicted scanpaths with $th=0.7$. (Second row) Four predicted scanpaths with $th=0.5$. (Third row) Four predicted scanpaths with $th=0.35$. (Fourth and fifth row) Predicted sequence of tSPM.}
\label{fig:i2300011983}
\end{figure*}

\begin{figure*}[t!]
\centering
\includegraphics[width=\linewidth]{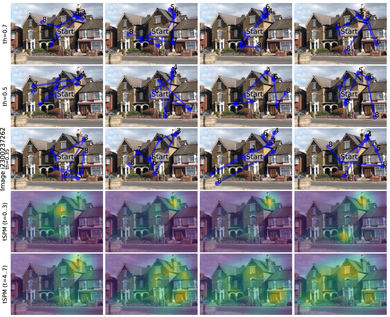}
\caption{Additional results for image \emph{i2300237262}. (First row) Four predicted scanpaths with $th=0.7$. (Second row) Four predicted scanpaths with $th=0.5$. (Third row) Four predicted scanpaths with $th=0.35$. (Fourth and fifth row) Predicted sequence of tSPM.}
\label{fig:i2300237262}
\end{figure*}

\begin{figure*}[t!]
\centering
\includegraphics[width=\linewidth]{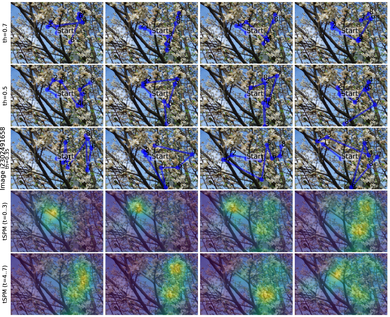}
\caption{Additional results for image \emph{i2302491658}. (First row) Four predicted scanpaths with $th=0.7$. (Second row) Four predicted scanpaths with $th=0.5$. (Third row) Four predicted scanpaths with $th=0.35$. (Fourth and fifth row) Predicted sequence of tSPM.}
\label{fig:i2302491658}
\end{figure*}

\begin{figure*}[t!]
\centering
\includegraphics[width=\linewidth]{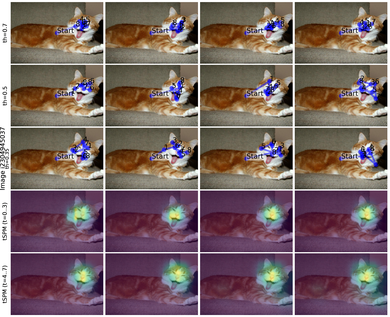}
\caption{Additional results for image \emph{i2304945037}. (First row) Four predicted scanpaths with $th=0.7$. (Second row) Four predicted scanpaths with $th=0.5$. (Third row) Four predicted scanpaths with $th=0.35$. (Fourth and fifth row) Predicted sequence of tSPM.}
\label{fig:i2304945037}
\end{figure*}

\begin{figure*}[t!]
\centering
\includegraphics[width=\linewidth]{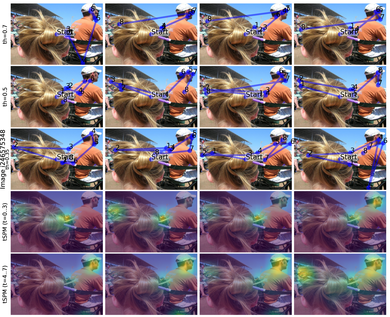}
\caption{Additional results for image \emph{i246575348}. (First row) Four predicted scanpaths with $th=0.7$. (Second row) Four predicted scanpaths with $th=0.5$. (Third row) Four predicted scanpaths with $th=0.35$. (Fourth and fifth row) Predicted sequence of tSPM.}
\label{fig:i246575348}
\end{figure*}

\begin{figure*}[t!]
\centering
\includegraphics[width=\linewidth]{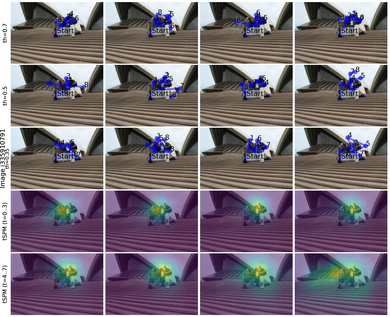}
\caption{Additional results for image \emph{i335910791}. (First row) Four predicted scanpaths with $th=0.7$. (Second row) Four predicted scanpaths with $th=0.5$. (Third row) Four predicted scanpaths with $th=0.35$. (Fourth and fifth row) Predicted sequence of tSPM.}
\label{fig:i335910791}
\end{figure*}

\end{document}

%% file: sections/RelatedWork.tex
\section{Related Work}
\label{sec:related}

\subsection{Saliency prediction} First approaches towards modeling human attention were based on saliency, as a measure of how much each part of a scene attracts human attention. The seminal work by Itti et al.~\cite{itti1998model} established the basis of visual attention prediction in images, by extracting hand-crafted features to generate a saliency map. This work inspired many posterior approaches (e.g., ~\cite{saliencytoolbox, zhao2011saliency}) which were also based on the computation of \textit{conspicuity maps} for different visual features (such as color, intensity, orientation of edges, or faces), which were then combined into a final saliency map. Other approaches included multiple semantic segmentation and surroundness analysis~\cite{lu2012cvpr}, or known human priors such as center bias or horizon line detectors~\cite{borji2012cvpr}, to improve saliency prediction.

With the proliferation of deep learning techniques and the appearance of public datasets~\cite{judd2009learning, mit-saliency-benchmark, yang2013saliency}, data-driven methods emerged, yielding impressive results. These methods were mostly based on convolutional neural networks (CNN) that extract latent features from which to infer saliency~\cite{Vig_2014_CVPR, kummerer2016deepgaze, Pan_2016_CVPR, martin20saliency}. Other approaches also leveraged the advances of generative networks~\cite{Pan_2017_SalGAN, xia2019predicting} and recurrent neural networks~\cite{cornia2018predicting, Wang_2018_CVPR}.
None of these works, however, take into account the dynamic nature of gaze behavior, not being able to model the temporal dimension of human attention.

\subsection{Scanpath prediction}

Scanpath models usually aim to progressively build a scanpath by concatenating single-point predictions, which may be partially based on the previous points of the path. Ellis and Smith~\cite{ellis1985patterns} presented a framework based on Markov stochastic processes. Later, other works proposed approaches that included known human biases, (such as the center bias, or human oculomotor constraints)~\cite{lemeur2015saccadic, tatler2009prominence,liu2013semantically,tavakoli2013stochastic}. Data-driven methods provide faster and more precise approaches, for instance using existing saliency prediction methods as a proxy to scanpath prediction, by means of winner-takes-all and inhibition-of-return strategies, by sampling heuristics~\cite{assens2017saltinet}, or simply leveraging deep features from neural networks~\cite{kummerer2016deepgaze}.

Scanpath prediction methods can be roughly categorized into (i) biologically inspired, (ii) statistically inspired, (iii) cognitively inspired, and (iv) engineered models~\cite{kummerer2021state}. Biologically inspired models take into account the importance of low-level features~\cite{itti1998model, tatler2017latest, zanca2019gravitational}, visual working memory~\cite{wang2011simulating}, attention and inhibition-of-return~\cite{engbert2015spatial}, or neuropsychology~\cite{adeli2017model}. Statistically inspired models try to mimic certain statistical properties of human scanpaths~\cite{boccignone2004modelling, sun2014toward, le2016introducing, clarke2017saccadic, xia2019predicting}. Cognitively inspired models assume that other cognitive processes besides low-level features can drive observers' attention, and therefore implement different human mechanisms such as low-level saliency, semantic and spatial effects~\cite{liu2013semantically} or region-of-interest and inhibition-of-return~\cite{sun2019visual}. Finally, engineered models just exploit the ability of data-driven techniques to fit to given data~\cite{chen2021predicting, assens2017saltinet, assens2018pathgan, bao2020human, hu2020dgaze}.

With this surge of data-driven approaches, and motivated by the temporal dependencies that human viewing behavior presents, some works have resorted to recurrent neural networks (RNN), which are capable of encoding previous information, and leveraging it to formulate a prediction~\cite{nguyen2018your}. However, scanpath prediction requires handling temporal and spatial information. To account for both, some recent approaches have built their models following ConvLSTM strategies~\cite{qiao2020viewport, li2019very, sun2019visual, xu2021spherical}, where convolutional operators handle spatial features while LSTM architectures enable learning temporal information.

However, all the aforementioned works are trained to optimize \textit{single-point} predictions, by means of direct losses such as MSE~\cite{assens2018pathgan} or BCE~\cite{sun2019visual}, and thus do not concern themselves with the plausibility of the scanpath as a whole. Recently, the work of Martin et al.~\cite{martin2021scangan360} presented a scanpath generation method for 360{$^\circ$} content, where the model was optimized by means of a dynamic time warping loss function on the whole distribution of ground-truth scanpaths, rather than on a single-point solution, and was hence able to learn and mimic latent behaviors in its predictions. 

In this work, and endorsed by previous literature, we resort to convolutional recurrent networks, but overcome the limitations of single-point prediction approaches by combining a novel loss function that combines dynamic time warping and Kullback-Leibler divergence, and a probabilistic approach. The loss function enables focusing on both the temporal and spatial aspects of the whole scanpaths, and optimizes our model over the whole distribution of real scanpaths, while our probabilistic approach accounts for the inherent human variability.

%% file: sections/Model.tex
\section{Our Model}
\label{sec:model}

 Our model performs probabilistic scanpath prediction given a single 2D image as input. The model, based on recurrent neural networks, is described in detail in this section:
we introduce the representation we employ for the scanpaths (Section~\ref{subsec:Model-ScanpathParameterization}), a novel loss function that is able to optimize our scanpaths in a joint spatio-temporal fashion (Section~\ref{subsec:Model-LossFunction}), our model architecture in depth (Section~\ref{subsec:Model-OurModel}), and additional details on our training data and procedure (Section~\ref{subsec:Model-TrainingDetails}).

\begin{figure}[t!]
  \centering
  \includegraphics[width=\linewidth]{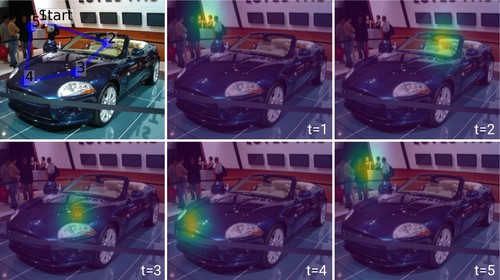}
  \caption{An example of our spatialized scanpath representation. The top left image shows a RGB image with a sample ground-truth scanpath overlaid. 
  For that sample ground-truth scanpath, we transform each fixation point into a Gaussian map centered at that particular point (see Section~\ref{subsec:Model-ScanpathParameterization}).}
  \label{fig:ExampleGaussianRepresentation}
\end{figure}

\begin{figure*}[th!]
  \centering
  \includegraphics[width=\linewidth]{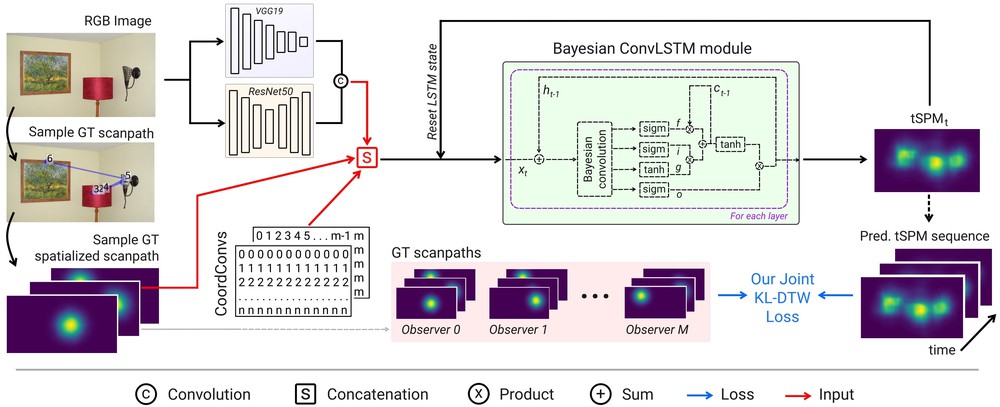}
  \caption{Overview of our model. We represent our scanpaths as Gaussian maps to enhance spatial learning (see Section~\ref{subsec:Model-ScanpathParameterization}). We leverage a pretrained VGG19 and a pretrained semantic segmentator based on ResNet50 to extract meaningful visual features from the image. We then combine those features and the spatialized scanpath with a CoordConv layer, to facilitate the learning of spatial features. This input is fed to a multi-layer convolutional LSTM module that iterates over the scanpath and predicts the next fixation, 
  until a scanpath of the target length is generated. We compute a loss based on dynamic time warping and Kullback-Leibler divergence (see Section~\ref{subsec:Model-LossFunction}) to optimize our model in a joint spatio-temporal fashion. The outcome of our model is a tSPM sequence (see also Figure~\ref{fig:TemporalEvolution}).}
  \label{fig:OurModel}
\end{figure*}

\subsection{Scanpath Representation}
\label{subsec:Model-ScanpathParameterization}

Scanpaths are commonly defined as a sequence $s = \{s_0, s_1, .., s_{N-1}\}$ of gaze points\footnote{In our case, and following common practice~\cite{bao2020human, sun2019visual, kummerer2016deepgaze}, the points in a scanpath correspond to fixation points (i.e., we do not attempt to predict saccades and other ocular movements).}, where $s_i = (x_i,y_i)$, and $(x_i,y_i)$ are the image coordinates of that particular gaze point.  
While this representation may suffice in some cases~\cite{fan2017fixation, zemblys2019gazenet, assens2018pathgan, martin2021scangan360}, it usually falls short for problems where it is necessary to establish a relationship between those coordinates and the position of features within an image. Indeed, convolutional networks are trained to be shift-invariant~\cite{liu2018coordconv}, and forcing them to explicitly learn the relation between gaze point coordinates and the actual positions of image features is challenging and hinders the training process.

Additionally, scanpaths for a given image exhibit both inter- and intra-observer variability. Not all observers will explore the image in exactly the same way, resulting in inter-observer variability. Besides, an observer watching the same image twice may follow slightly different scanpaths, and, even if asked to follow a certain path, there is a ballistic or noisy component in ocular movements (e.g., saccades or post-saccadic oscillations~\cite{larsson2013detection}), that results in different gaze points. As a result, scanpaths are non-deterministic. However, they do exhibit clear patterns across and within observers, as multiple works have shown~\cite{lemeur2015saccadic}. 

Given this variability, and in order to facilitate the spatial learning of the network,  %
instead of representing each gaze point with its coordinates, $s_i = (x_i,y_i)$,
a more adequate representation for $s_i$ is a Gaussian distribution $g^s_i$ centered in $(x_i,y_i)$, and defined over the whole image. In each distribution $g_i$, there is thus a value $g_i(x,y)$ per pixel $(x,y)$, which represents the probability of a gaze point falling at pixel $(x,y)$ at time step $i$. A scanpath $s$ is therefore represented as a sequence $g^s = \{g^s_0, g^s_1, .., g^s_{N-1}\}$ of Gaussian maps $g^s_i$ (see Figure~\ref{fig:ExampleGaussianRepresentation}); we term a scanpath represented in this way a \emph{spatialized} scanpath. This representation facilitates spatial learning by providing a direct correlation between a scanpath and its corresponding image.

\subsection{Overview of the Model}
\label{subsec:Model-Overview}

Our model, illustrated in Figure~\ref{fig:OurModel}, is based on the recently presented ConvLSTM~\cite{xingjian2015convolutional}, a type of recurrent neural network. %
ConvLSTMs maintain the recurrent nature of traditional LSTMs, processing data in a sequential manner, thus being able to learn the temporal features of the data. Additionally, ConvLSTMs are provided with convolutional operators that handle visual information and facilitate learning spatial features in the input sequence. 

\begin{figure}[t!]
  \centering
  \includegraphics[width=\linewidth]{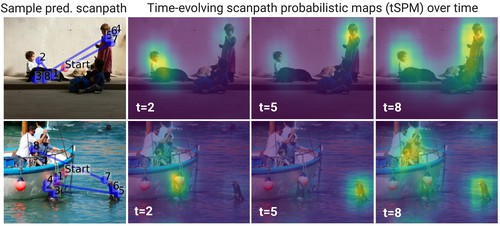}
  \caption{We show, for two given test images, the evolution of our predicted time-evolving scanpath probabilistic maps (tSPM, see Section~\ref{subsec:Model-Overview}) over time. Our Bayesian approach enables 
  a probabilistic weight selection for the next fixation, and allows our model to be stochastic, while the ConvLSTM module allows taking into account the previous fixations (see Section~\ref{subsec:Model-OurModel}). }
  \label{fig:TemporalEvolution}
\end{figure}

Further, we resort to a Bayesian approach when modeling the ConvLSTM module, %
in order to better incorporate the uncertainty driven by inter- and intra-observer variability: The output of the ConvLSTM module is not a point, but rather a probability map (see Figure~\ref{fig:OurModel}). Our whole model therefore predicts, given an input image, a sequence of time-evolving scanpath probabilistic maps (tSPM) (see Figure~\ref{fig:TemporalEvolution}). Each tSPM represents the probabilities of the next gaze fixation point falling on each pixel of the image at a certain time instant.

We additionally leverage pretrained neural networks on image classification tasks to facilitate feature extraction, and CoordConv layers to improve learning of spatial features. The details of our model architecture are described in Section~\ref{subsec:Model-OurModel}.

\subsection{Loss Function}
\label{subsec:Model-LossFunction}

Our spatialized scanpath representation facilitates working over the spatial component of the scanpaths, as explained in Section~\ref{subsec:Model-ScanpathParameterization}. Both the spatial and temporal domains are critical when predicting gaze points. Recurrent neural networks (RNN) have proven to be powerful tools able to handle time dependencies in data, being able to extract, maintain and even infer patterns through time, and thus have been successfully used in some approaches for gaze prediction~(see Section~\ref{sec:related}). However, all those approaches have designed their RNN-based models to optimize the prediction at each time step, with element-wise loss functions, such as mean squared error (MSE) or binary cross-entropy (BCE), that penalize the prediction for \emph{each point} in isolation. %

In contrast, we propose a novel loss function based on the Kullback-Leibler divergence and dynamic time warping, computed over the whole scanpath. The former allows our model to account for the spatial relations between gaze points, while the latter ensures a realistic and plausible temporal behavior of the predicted scanpaths. 

\bigbreak
 \emph{Kullback-Leibler Divergence (KL-Div)} KL-Div is a measure of how different a probability distribution is from another one, and is one of the most commonly used metrics and losses in saliency prediction problems~\cite{zhang2020spatial, qiao2020viewport, palazzi2018predicting, wu2020salsac}. The Kullback-Leibler divergence ($D_{KL}$) is defined as:

\begin{equation}
    D_{KL} (P||Q) = \sum_{j} P(j) ln\frac{P(j)}{Q(j)} ,
\label{eq:KL}
\end{equation}

\noindent where $P$ and $Q$ are the probability distributions to be compared, and $j$ refers to each point
of the distribution. In our particular case, each gaze point is represented in a spatialized manner, hence KL-Div is able to give a qualitative measurement on how different two points are based on their probability distributions $P$ and $Q$.

\bigbreak
\emph{Dynamic Time Warping (DTW)} DTW is a measure of similarity between two time series that may differ in length or speed~\cite{muller2007dynamic}. The DTW algorithm attempts to find the optimal match between the points of two temporal sequences, $r$ and $s$, by matching each point in one of them with at least one point in the other, without forcing a one-to-one correspondence between both sequences. The optimal match is found by minimizing a cost function: a distance matrix $\Delta$ stores the cost (Euclidean distance) for each possible pair of points, and the optimization searches for the matching (alignment) between $r$ and $s$ such that the total cost is minimized. This can be written as:

\begin{equation}
\label{eq:dtw_loss_1}
DTW(r, s) = \underset{A}{\min}
 \langle A, \Delta(r, s) \rangle, 
\end{equation}

\noindent where $A$ is a binary alignment matrix between two time series $r$ and $s$, $\Delta(r,s) = [\delta(r_i,s_j)]_{i,j}$ is a matrix containing the distances $\delta(\cdot , \cdot )$ between each pair of points in $r$ and $s$, and $\langle \cdot , \cdot \rangle$ denotes the inner product between both matrices.
Since the minimum function is not differentiable, a soft version has been proposed~\cite{cuturi2017soft}: 

\begin{equation}
\label{eq:dtw_loss}
DTW^{\gamma}(r, s) = \underset{A}{\min}^{\gamma}
 \langle A, \Delta(r, s) \rangle, \quad \gamma > 0
\end{equation}

\noindent The soft-min function $\min^{\gamma}$ is defined as: 

\begin{equation}
\label{eq:dtw_loss_min}
{\min}^{\gamma}(a_1,...,a_N) = -\gamma\:log\sum_{i=1}^N exp\left(-\frac{a_i}{\gamma}\right) ,
\end{equation}

\noindent with the $\gamma$ parameter adjusting the similarity between the soft version and the original DTW algorithm, both being the same when $\gamma = 0$. 
Eq.~\ref{eq:dtw_loss} has been used successfully as a loss term in related contexts, such as scanpath generation for virtual reality~\cite{martin2021scangan360} or weakly supervised action alignment and segmentation in videos~\cite{Chang_2019_CVPR}.

\bigbreak
\emph{Our Joint KL-DTW Loss} %
While KL-Div accounts for the spatial similarity of two distributions, and DTW focuses on the temporal dimension,
none of them suffices on its own in our particular case. We therefore propose a novel loss function based on the combination of both KL-Div and DTW, defined as follows:

\begin{equation}
    \mathcal{L}_{KL-DTW}(r') = \frac{\sum\limits_{s=1}^S DTW^{\gamma}(r',s)}{|S|} ,
\end{equation}

\noindent where $r'$ is a predicted sequence of tSPM (see Section~\ref{subsec:Model-Overview}), and $s$ is a  ground-truth scanpath from the set of ground-truth ones $S$ for a given image $I$. $DTW^{\gamma}$ is computed as given by Eq.~\ref{eq:dtw_loss}. However, we modify the computation of the distance matrix $\Delta(r',s) = [\delta(r'_i,s_j)]_{i,j}$ such that, instead of $\delta$ being an Euclidean distance, we have: 

\begin{equation}
    \delta(r'_i,s_j) =  D_{KL} (r'_i||g^s_j),
\label{eq:delta_maps}
\end{equation}

\noindent  where $r'_i$ is the $i^{th}$ predicted tSPM, $g^s_j$ is the spatialized representation of point $s_j$ as described in Section~\ref{subsec:Model-ScanpathParameterization}, and $D_{KL}$ is the Kullback-Leibler divergence (Eq.~\ref{eq:KL}).

This formulation allows our model to be optimized to find an alignment that minimizes both the \emph{spatial} and the \emph{temporal} differences between each predicted scanpath and the ground-truth ones, therefore predicting scanpaths that follow a similar distribution as the ground truth. To our knowledge, we are the first to propose such a combination of metrics.

\bigbreak
\emph{Bias Regularization Term} Human gaze data in 2D images is known to be strongly biased towards the center of the images~\cite{lemeur2015saccadic}. Although inherent to human nature, such bias hinders the learning process of the network, which can easily overfit to that behavior. Based on this, we include a regularization loss term that penalizes scanpaths whose points tend to stay in the center of the image for a long time, hence eliciting a more exploratory behavior that better reflects ground truth data. Our regularization term is included in the pairwise cost computations $\delta(r'_i,s_j)$ for the distance matrix $\Delta$, modifying Eq.~\ref{eq:delta_maps} as follows: %
\begin{equation}
    \delta(r'_i,s_j) =  D_{KL} (r'_i||g^s_j) + \lambda_{CB} * \mathcal{L}_{Reg}(r'_i) 
\label{eq:delta_wReg}
\end{equation} 
\begin{equation}
    \mathcal{L}_{Reg}(r'_i) = \frac{1}{D_{KL}(r'_i||g_c)} ,
\label{eq:Reg}
\end{equation}
where 
$g_c$ is a Gaussian map representing the aforementioned center bias, computed following the representation introduced in Section~\ref{subsec:Model-ScanpathParameterization} for a point $c$ in the center of the image.

\begin{figure}[t!]
  \centering
  \includegraphics[width=\linewidth]{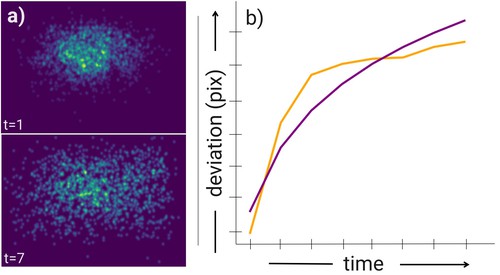}
  \caption{When observing images, there is a strong bias towards the center of the image~\cite{lemeur2015saccadic}. This bias can hinder the learning process of deep networks, with a significant risk of overfitting to that behavior. We have analyzed that bias in the datasets used in this work (Section~\ref{subsec:Model-TrainingDetails}), and found that this behavior diminishes over time. (a) Distribution of scanpath fixations in two different time instants for the whole training dataset. (b) We have computed the standard deviations of fixations (y-axis) over time (x-axis) for all the ground-truth scanpaths (in orange), and found they increase in a logarithmic fashion (the fitted curve is shown in purple). We introduce a regularization term in our loss function to foster a similar behavior in our predicted scanpaths (see Section~\ref{subsec:Model-LossFunction}).}
  \label{fig:Bias}
\end{figure}

In order to set the relative weight of the regularization term $\lambda_{CB}$, we analyzed the datasets used (see Section~\ref{subsec:Model-TrainingDetails}), and found that this center bias behavior diminishes over time, with fixations being more widely spread over the image in later time instants. We measured the standard deviation of fixation positions in the ground-truth data, and found them to increase logarithmically over time ($R^2 = 0.855$); we increase $\lambda_{CB}$ in the same way (see Figure~\ref{fig:Bias}).

\begin{figure*}[t!]
  \centering
  \includegraphics[width=\linewidth]{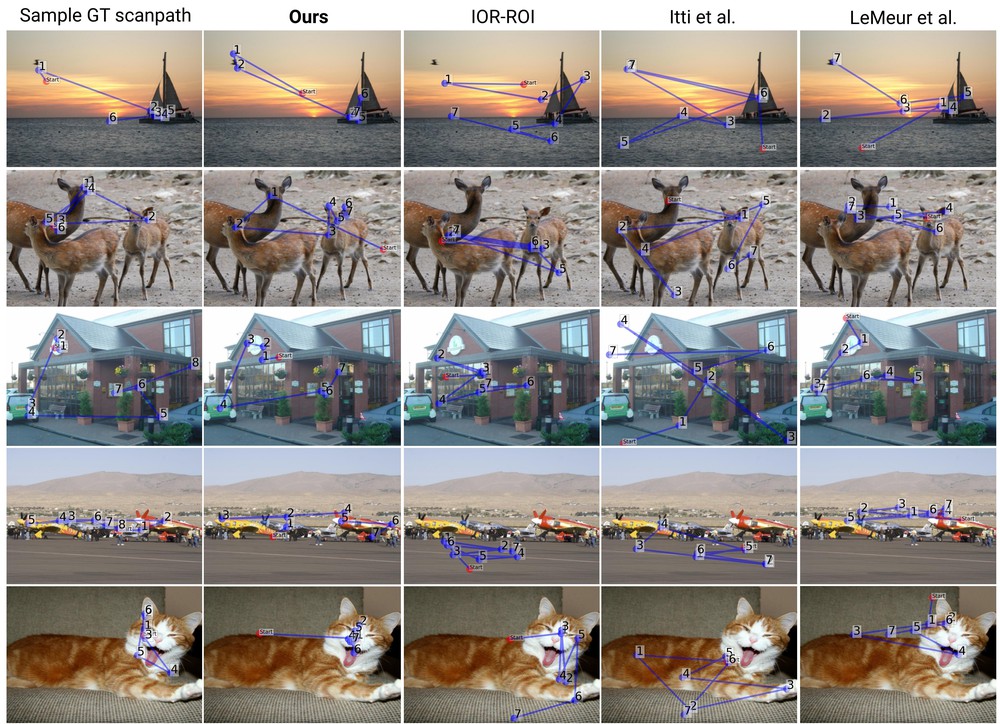}
  \caption{Comparison to other approaches for scanpath prediction. Each row shows representative scanpaths for a certain image: ground truth, our method, the IOR-ROI model from Sun et al.~\cite{sun2019visual}, Itti et al.~\cite{itti1998model}, and LeMeur et al.~\cite{lemeur2015saccadic}. Both Itti et al.'s and Sun et al.'s methods sometimes fail to fixate in salient regions, leading to unnatural scanpaths. Different from LeMeur's method, our model is able to focus on the regions of interest while maintaining a plausible exploration trajectory. %
  Note that this figure only contains one sample scanpath per method for illustrative purposes. A more thorough analysis with quantitative metrics can be found in Table~\ref{tab:ComparisonMetrics}, and additional qualitative results can be found in the supplementary material.}
  \label{fig:GridComparison}
\end{figure*}

\subsection{Model Architecture}
\label{subsec:Model-OurModel}

Our model features a recurrent neural network (RNN), which is able to extract and maintain temporal latent information from the scanpaths it is trained with. Particularly, we choose a convolutional long short-term memory (ConvLSTM) network~\cite{xingjian2015convolutional}, which is an adaptation of classic LSTMs to work with 2D data, such as images. This type of network has proven to be effective in many different problems, such as weather forecasting~\cite{xingjian2015convolutional}, video saliency detection~\cite{song2018pyramid}, or medical image segmentation~\cite{azad2019bi}. 

ConvLSTMs behave in a similar way to traditional LSTMs, working over four different gates; however, since they handle spatial data, they conduct convolutional operations rather than lineal ones. The ConvLSTM used in this work\footnote{https://github.com/ndrplz/ConvLSTM\_pytorch} is defined as follows:

\begin{gather}
    i_t = \sigma(Conv(x_t;w_{xi}) + Conv(h_{t-1};w_{hi}) + b_i)\nonumber\\
    f_t = \sigma(Conv(x_t;w_{xf}) + Conv(h_{t-1};w_{hf}) + b_f)\nonumber\\
    o_t = \sigma(Conv(x_t;w_{xo}) + Conv(h_{t-1};w_{ho}) + b_o)\nonumber\\
    g_t = Tanh(Conv(x_t;w_{xg}) + Conv(h_{t-1};w_{hg}) + b_g)\nonumber\\
    c_t = f_t \odot c_{t-1} + i_t \odot g_t\nonumber\\
    h_t = o_t \odot Tanh(c_t)
\label{eq:ConvLSTM}
\end{gather}

\noindent where $x_t$ is the input at a time step $t$, and $h_{t}$ is the hidden state of the network up to the current time step, which also serves as the output of the network. We refer the reader to the work of Xingjian et al.~\cite{xingjian2015convolutional} for an in-depth explanation of the ConvLSTM architecture.

In our particular case of scanpath prediction, there is a degree of stochasticity driven by the inter- and intra-observer variability. As a result, given a particular trajectory (sequence of gaze fixation points), instead of predicting the next point in a deterministic manner, we predict a probability distribution (i.e., the previously introduced tSPM). %
Due to this, we combine the aforementioned ConvLSTM with the recently introduced Bayesian deep learning~\cite{wang2016towards, kendall2017uncertainties}. Unlike traditional deep learning (DL), where the weights of the network are deterministic, in Bayesian DL the weights of a particular layer are sampled from a probability distribution, and thus the network itself can account for the inherent uncertainty of the data~\cite{kendall2017uncertainties}. During the network training, those weight distributions are also optimized. Thus, in our case, we substitute each convolutional operation from Eq.\ref{eq:ConvLSTM} with a Bayesian 2D convolution. This way, our ConvLSTM will no longer be deterministic, and will be able to account for the stochastic component of scanpaths.

\begin{table*}[t!]
\centering
\caption{Results of our quantitative comparisons. We first include upper (human baseline, \emph{Human BL}) and lower (random scanpaths, \emph{Random BL}) baselines for reference. Then, we compare to Sun et al.'s IOR-ROI~\cite{sun2019visual}, Itti et al.~\cite{itti1998model}, and LeMeur et al.~\cite{le2016introducing}. We also compute our set of metrics varying our probabilistic threshold $th$ (see Section~\ref{subsec:Model-OurModel}). Arrows indicate whether higher or lower is better; boldface highlights the best result for each metric. Overall our model ($th=0.7$) yields the best performance across metrics, closest to the human baseline. Moreover, as the last two lines show, decreasing the value of our parameter $th$ (thus leading to more variability in our scanpaths), still leads to good results (see Section~\ref{subsec:EvalScanpathVar} for details).}
\label{tab:ComparisonMetrics}
\arrayrulecolor{black}
\resizebox{\linewidth}{!}{%
\begin{tabular}{c!{\color{black}\vrule}cc|cc|cc|cccc}
\multicolumn{1}{l!{\color{black}\vrule}}{} & \multicolumn{2}{c|}{String alignment}        & \multicolumn{2}{c|}{Curve similarities}           & \multicolumn{2}{c|}{Time-series analysis}        & \multicolumn{4}{c}{Recurrence analysis}                                                     \\
Model                                      & LEV$\downarrow$                   & SCAM$\uparrow$                  & HAU$\downarrow$                      & FRE$\downarrow$                      & fDTW$\downarrow$                      & TDE$\downarrow$                    & REC$\uparrow$                  & DET$\uparrow$                  & LAM$\uparrow$                  & CORM$\uparrow$                   \\ 
\arrayrulecolor{black}\hline
Human BL                                   & 10.77 (1.61)          & 0.38 (0.06)          & 95.97 (18.40)           & 140.02 (26.16)          & 550.84 (133.71)          & 42.40 (8.45)          & 6.69 (3.74)          & 1.72 (1.51)          & 6.09 (6.01)          & 22.11 (7.41)           \\
Random BL                                  & 12.31 (0.88)          & 0.20 (0.02)          & 148.01 (13.76)          & 199.30 (13.63)          & 877.15 (71.66)           & 69.87 (4.34)          & 0.73 (0.34)          & 0.02 (0.09)          & 0.19 (0.25)          & 3.79 (1.74)            \\ 
\arrayrulecolor[rgb]{0.753,0.753,0.753}\hline
Ours                                       & \textbf{11.47 (1.13)} & 0.34 (0.06)  & \textbf{103.44 (27.13)}  & \textbf{144.77 (32.77)} & \textbf{610.02 (155.96)} & 43.74 (10.25) & \textbf{3.52 (2.86)} & \textbf{0.64 (0.84)} & \textbf{5.05 (4.96)} & \textbf{13.95 (7.92)}  \\
IOR-ROI                                    & 13.26 (0.71)          & 0.30 (0.05)          & 115.50 (20.22)          & 166.07 (21.69)          & 777.75 (119.46)          & 46.98 (7.18)          & 1.80 (0.98)          & 0.18 (0.31)          & 0.81 (1.35)          & 10.28 (4.43)           \\
Itti et al.                                & 14.04 (0.80)          & 0.23 (0.05)          & 160.09 (29.31)          & 207.97 (27.21)          & 1041.16 (153.97)         & 63.88 (9.54)          & 1.02 (1.98)          & 0.04 (0.22)          & 0.62 (2.03)          & 5.84 (6.00)            \\
LeMeur et al.                              & 12.58 (0.78)  & \textbf{0.35 (0.04)} & 104.84 (12.79) & 163.59 (20.52)  & 669.67 (108.49)  & 39.75 (6.53) & 2.39 (1.18)  & 0.40 (0.48)  & 2.09 (2.26)  & 12.54 (4.45)\\
\arrayrulecolor[rgb]{0.753,0.753,0.753}\hline
Ours (th = 0.5)                             & 11.60 (0.98) & 0.33 (0.06)            & 103.97 (23.23) & 149.37 (30.09)         & 636.08 (146.44) & 45.46 (10.30)           & 3.01 (2.28) & 0.50 (0.58) & 3.14 (2.85) & 12.96 (6.73)  \\ 
\cline{7-7}
Ours (th = 0.35)                             & 13.26 (0.71) & 0.30 (0.05)            & 102.17 (19.77) & 149.99 (28.42)         & 639.07 (138.27) & 45.77 (9.63)            & 2.82 (2.09) & 0.44 (0.54) & 2.43 (2.59) & 12.88 (6.21)  \\
\end{tabular}%
}
\arrayrulecolor{black}
\end{table*}

Instead of feeding our Bayesian ConvLSTM with the raw images, we preprocess them to facilitate the learning process and enhance our model's performance. For this, we first extract the main image features with a pretrained VGG\-19~\cite{russakovsky2015imagenet, simonyan2014very}, and a semantic segmentation mask with a pretrained ResNet50~\cite{he2016deep}. We then convolve both of them together, to obtain a final, comprehensive, single-channel image feature representation. At each time step, this feature representation is fed to the ConvLSTM alongside with (i) the corresponding Gaussian map representing a fixation, and (ii) a CoordConv layer~\cite{liu2018coordconv}. CoordConv layers have proven to ease spatial learning and facilitate network convergence. Different from previous approaches~\cite{itti1998model, le2016introducing, sun2019visual}, we do not resort to saliency for scanpath prediction, since saliency is an aggregated spatial notion that has lost the temporal information, and is not usually available as ground truth.

With this input, our network is able to predict a tSPM (see Section~\ref{subsec:Model-Overview}). Then, to choose the actual pixel where the gaze point will fall, we follow a probabilistic weighted sampling strategy, that again accounts for the stochastic nature of human visual exploration. We first discard all pixels with a probability lower than a threshold $th = 0.7$ (see Section~\ref{subsec:EvalScanpathVar} for additional evaluation on this), and then sample the next point based on the predicted map's probabilities. Once a point $s_i = (x_i,y_i)$ has been sampled, a Gaussian map centered in $(x_i,y_i)$ is again computed (see Section~\ref{subsec:Model-ScanpathParameterization}), and fed to the network for its posterior predictions, until the whole scanpath is predicted. 

Once the whole scanpath has been predicted, our novel KL-DTW loss (see Section~\ref{subsec:Model-LossFunction}) is computed between the ground truth and the sequence of tSPM, and the network is then optimized. A complete overview of our model can be seen in Figure~\ref{fig:OurModel}.

\subsection{Datasets and Training Details}
\label{subsec:Model-TrainingDetails}

Following previous work~\cite{sun2019visual}, we train our model over the OSIE dataset~\cite{xu2014predicting}, which contains 700 different images with their corresponding gaze information for a total of fifteen observers, yielding a total of approximately 10,500 scanpaths. We again follow previous work~\cite{sun2019visual} and discard all the scanpaths with $N < 4$, %
and generate scanpaths of length $N = 8$, which is the mean length of our ground-truth data. 

For the rest of the scanpaths, in order to train our model, we preprocess each to follow the representaton introduced in Section~\ref{subsec:Model-ScanpathParameterization}. To validate our model, and again inspired by previous approaches~\cite{sun2019visual}, we use the MIT low resolution dataset~\cite{judd2011fixations}. Please refer to Section~\ref{sec:evaluation} for further details. 

We trained our model using the Hydra~\cite{Yadan2019Hydra} and Pytorch Lightning~\cite{PytorchLightning} frameworks for PyTorch, logging and checkpointing all the necessary parameters to restore the training process at any point. We use the Adam optimization algorithm~\cite{Adam}. The learning rate has a value of $10^{-4}$, and we set batch size to 1. We trained our model on a Nvidia RTX 2080 Ti with 11GB of VRAM until convergence, for a total of 22 hours.

%% file: sections/Evaluation.tex
\section{Evaluation}
\label{sec:evaluation}

We validate the quality of our scanpaths against measured, ground-truth scanpaths, as well as to other existing scanpath prediction methods. 
Similar to recent work on scanpath generation~\cite{martin2021scangan360}, we rely on the comprehensive set of metrics proposed by Fahimi and Bruce~\cite{fahimi2020metrics}, which include string alignment, curve similarity, time-series analysis, and recurrence analysis. We refer the reader to the original publication for further details on the metrics. In addition, we also analyze the performance of our method against ground-truth data for spatial convergence and saliency, inter-observer variability, and fixation prediction.

\begin{figure}[t!]
    \centering
    \includegraphics[width=\linewidth]{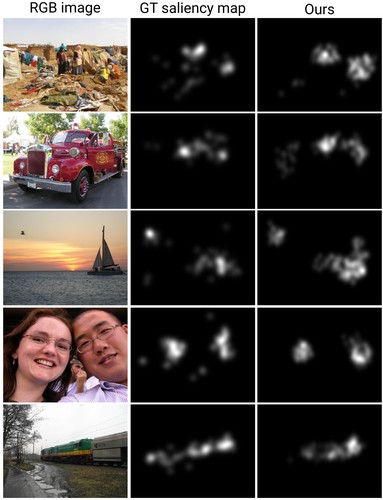}
    \caption{We evaluate the spatial convergence of our predicted scanpaths: Our model generates scanpaths that focus on salient regions, and whose aggregation closely resembles ground-truth saliency maps.}%
    \label{fig:ConvSaliency}
\end{figure}

\begin{figure*}[t!]
    \centering
    \includegraphics[width=\linewidth]{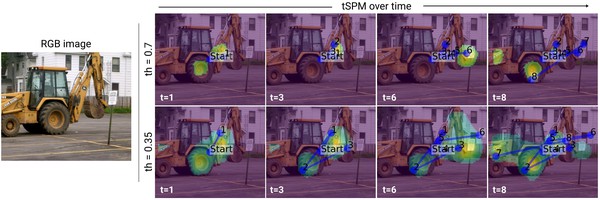}
    \caption{We evaluate the ability of our model to generate scanpaths with higher variability. Our model has a stochastic nature, and generates scanpaths by probabilistically sampling the predicted tSPM (see Section~\ref{subsec:Model-OurModel}). Any point with a probability below a specified threshold $th$ will be discarded before sampling. By lowering this value, points with lower probability can be sampled, leading to scanpaths that can focus on regions further from the main regions of interest, yielding a more exploratory behavior. Nevertheless, even when reducing this threshold, our scanpaths remain stable in quantitative evaluations (see the last two rows of Table~\ref{tab:ComparisonMetrics}).}
    \label{fig:DiffThresholds}
\end{figure*}

\subsection{Comparison to Other Approaches}
\label{subsec:EvalComparison}

Following previous literature in scanpath prediction~\cite{sun2019visual}, we generate ten scanpaths per image for our test set with each of the methods we are comparing against: Itti et al.'s ~\cite{itti1998model}, LeMeur et al.'s work~\cite{lemeur2015saccadic}, and IOR-ROI~\cite{sun2019visual}. Scanpath length is determined by the mean length of ground-truth data~\cite{sun2019visual}. An illustrative qualitative comparison can be seen in Figure~\ref{fig:GridComparison}. Since some of the models~\cite{itti1998model, sun2019visual} are based on biological mechanisms such as inhibition of return, fixations do not remain in the same region, leading to unnatural scanpaths. %
Our model and LeMeur et al.'s~\cite{lemeur2015saccadic} produce scanpaths that more closely resemble the ground truth. However, our work does not depend on saliency as a proxy, and does not require a module devoted to its prediction. This makes it more general and suitable for data for which ground-truth saliency is not available.

Table~\ref{tab:ComparisonMetrics} shows the comparisons with quantitative metrics~\cite{fahimi2020metrics}. For reference, we also include human baseline (Human BL)~\cite{xia2019predicting} by computing the same metrics for all the ground-truth scanpaths, plus a random baseline (Random BL) generated from random scanpaths. Our models yields the best results in eight of the ten cases, and second in the remaining two.

Additional qualitative results can be found in the supplementary material.

\subsection{Spatial convergence and saliency} 
\label{subsec:EvalSaliency}

To evaluate the spatial convergence of our predicted scanpaths, we compare saliency maps. We compute such maps by aggregating multiple scanpaths into a heatmap; we then compare them against the ground-truth saliency maps computed from real observers' data. As can be seen in Figure~\ref{fig:ConvSaliency}, our generated scanpaths lead to predicted saliency maps that closely resemble the ground truth.

\subsection{Scanpath variability}
\label{subsec:EvalScanpathVar}
We generate our scanpaths by sampling our %
generated tSPM (Section \ref{sec:model}). The inherent variability that exists between different observers is modeled by the parameter $th$: higher values lead to more concentrated scanpaths, while lower ones allow our scanpaths to simulate more exploratory visual behaviors.  Figure~\ref{fig:DiffThresholds} illustrates this. In addition, we have conducted a quantitative analysis (see last two rows of Table~\ref{tab:ComparisonMetrics}) showing how, even when eliciting a more exploratory behavior by decreasing $th$, our scanpaths still outperform previous approaches and remain close to the human baseline.

\subsection{Step-wise fixation prediction} 
\label{subsec:EvalStepWise}

As mentioned in Section~\ref{sec:related}, most existing works take an incomplete scanpath as input, and predict the \textit{next} fixation point. They thus build each scanpath progressively, usually by
optimizing only the prediction of that last point (e.g., by means of MAE~\cite{hu2020dgaze} or MSE~\cite{nguyen2018your} losses).
Although this approach neglects the plausibility of the full scanpath as a whole, it may be useful in some cases. Our proposed spatio-temporal loss and probabilistic framework  also offer a precise alternative in these situations. 
Table~\ref{tab:pointwise} shows quantitative results for paths of varying lengths: $\ocircle$ represents points from the ground-truth scanpath fed to our network, while $\times$ represents points predicted with our model. 
Our method produces plausible results from a single ground-truth point, and very quickly approximates the human baseline with only four.

\begin{table}
\centering
\caption{We have evaluated the ability of our model to complete sequences of scanpaths of different lengths: $\ocircle$ represents points from the ground-truth scanpath, and $\times$ points predicted with our model. Each row is computed by completing every scanpath on our test set 10 times (see Section~\ref{subsec:EvalStepWise}). The first and last rows show a random baseline and the human baseline, respectively.}
\label{tab:pointwise}
\arrayrulecolor{black}
\begin{tabular}{c|cccc}
 Scanpath         & SCAM$\uparrow$ & HAU$\downarrow$  & fDTW$\downarrow$  & REC$\uparrow$   \\ 
\hline
Random BL & 0.21 & 192.96 & 703.19 & 0.72  \\ 
\arrayrulecolor[rgb]{0.753,0.753,0.753}\hline
 $\ocircle \times \times \times \times$  & 0.42 & 131.02 & 421.72 & 5.16   \\
 $\ocircle \ocircle \times \times \times$  & 0.44  & 129.52 & 407.57 &  6.32 \\
 $\ocircle \ocircle \ocircle \times \times$  & 0.46 & 128.81  & 393.83 &  6.99  \\
 $\ocircle \ocircle \ocircle \ocircle \times$  & 0.47 & 129.03  & 381.93 & 7.24  \\ 
\hline
Human BL  & 0.49 & 126.73 & 387.75 & 7.17 
\end{tabular}
\arrayrulecolor{black}
\end{table}

%% file: sections/Conclusions.tex
\section{Conclusion}
\label{sec:Conclusion}

\begin{figure}[t!]
    \centering
    \includegraphics[width=\linewidth]{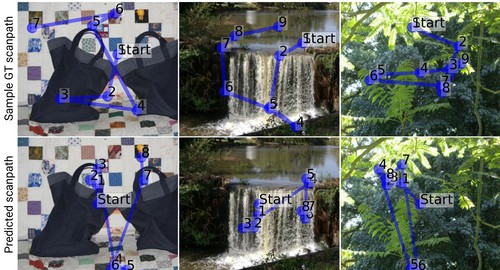}
    \caption{When the visual features of an image are complex or too abstract, our model's performance diminishes: It sometimes fails to localize the areas of interest, or remains within the same areas for several fixations. Feeding our model with larger datasets with more varied images would probably make it more robust to these cases.}
    \label{fig:Failure}
\end{figure}

We have presented a novel method for scanpath prediction in 2D images. We introduce a novel spatial scanpath representation that enhances learning the spatial features of images, together with a novel loss function tailored to the spatio-temporal particularities of scanpaths, based on a combination of dynamic time warping and Kullback-Leibler divergence. This allows our model to predict scanpaths that mimic human viewing patterns. We have evaluated our model and compared it to state-of-the-art methods on a large set of metrics that analyze different aspects of scanpaths. Our model outperforms previous approaches, while generating a scanpath in less than a second. 

\bigbreak
\emph{Limitations and future work} Our work is not free from limitations, and offers interesting avenues of future work. When the visual features of the image are too abstract or complex, the performance of our model decreases (see Figure~\ref{fig:Failure}). We computed the same set of metrics as in Section~\ref{sec:evaluation} and found that, for some particular complex cases, our metrics are closer to the random baseline than to the human baseline. 
We hypothesize that using a larger dataset and more ground-truth data would ameliorate this. In addition, adding more priors may be helpful, although finding out what priors would apply to the most complex cases is still an open problem. 

Additionally, exploring the impact of the \textit{duration} of the fixations could further enhance our model's performance. Last, our model assumes no prior knowledge or task-oriented scenarios when viewing the images; it would be interesting to devise variations of our model for such particular cases.